\documentclass[a4paper,fleqn]{cas-dc}

\usepackage[numbers]{natbib}

\def\tsc#1{\csdef{#1}{\textsc{\lowercase{#1}}\xspace}}
\tsc{WGM}
\tsc{QE}
\tsc{EP}
\tsc{PMS}
\tsc{BEC}
\tsc{DE}
\usepackage{wrapfig}
\usepackage[labelsep=colon]{caption}
\usepackage{subcaption}
\usepackage{algorithm,algpseudocode}
\usepackage{multirow}
\usepackage{csquotes}
\usepackage{balance} 

\algnewcommand{\algorithmicforeach}{\textbf{for each}}
\algdef{SE}[FOR]{ForEach}{EndForEach}[1]
  {\algorithmicforeach\ #1\ \algorithmicdo}
  {\algorithmicend\ \algorithmicforeach}

\begin{document}
\shorttitle{SpaceNet: Make Free Space For Continual Learning}

\title [mode = title]{SpaceNet: Make Free Space For Continual Learning}                      

\tnotetext[1]{Code available at: https://github.com/GhadaSokar/SpaceNet}


\author[1]{Ghada Sokar}
\cormark[1]

\address[1]{Department of Mathematics and Computer Science, Eindhoven University of Technology, The Netherlands}

\author[1,2]{Decebal Constantin Mocanu}

\author[1]{Mykola Pechenizkiy}

\address[2]{Faculty of Electrical Engineering, Mathematics, and Computer Science, University of Twente, The Netherlands}

\cortext[cor1]{Corresponding author}

\begin{abstract}
The continual learning (CL) paradigm aims to enable neural networks to learn tasks continually in a sequential fashion. The fundamental challenge in this learning paradigm is catastrophic forgetting previously learned tasks when the model is optimized for a new task, especially when their data is not accessible. Current architectural-based methods aim at alleviating the catastrophic forgetting problem but at the expense of expanding the capacity of the model. Regularization-based methods maintain a fixed model capacity; however, previous studies showed the huge performance degradation of these methods when the task identity is not available during inference (e.g. class incremental learning scenario).
In this work, we propose a novel architectural-based method referred as SpaceNet for class incremental learning scenario where we utilize the available fixed capacity of the model intelligently. SpaceNet trains sparse deep neural networks from scratch in an adaptive way that compresses the sparse connections of each task in a compact number of neurons. The adaptive training of the sparse connections results in sparse representations that reduce the interference between the tasks. Experimental results show the robustness of our proposed method against catastrophic forgetting old tasks and the efficiency of SpaceNet in utilizing the available capacity of the model, leaving space for more tasks to be learned. In particular, when SpaceNet is tested on the well-known benchmarks for CL: split MNIST, split Fashion-MNIST, CIFAR-10/100, and iCIFAR100, it outperforms regularization-based methods by a big performance gap. Moreover, it achieves better performance than architectural-based methods without model expansion and achieves comparable results with rehearsal-based methods, while offering a huge memory reduction.
\end{abstract}


\maketitle

\section{Introduction}
Deep neural networks (DNNs) have achieved outstanding performance in many computer vision and machine learning tasks \cite{he2015delving,zoph2018learning,chen2017deeplab,kenton2019bert,lin2017feature,guo2016deep,liu2017survey}. However, this remarkable success is achieved in a static learning paradigm where the model is trained using large training data of a specific task and deployed for testing on data with similar distribution to the training data. This paradigm contradicts the real dynamic world environment which changes very rapidly. Standard retraining of the neural network model on new data leads to significant performance degradation on previously learned knowledge, a phenomenon known as catastrophic forgetting \cite{mccloskey1989catastrophic}. Continual learning, also called as lifelong learning, comes to address this dynamic learning paradigm. It aims at building neural network models capable of learning sequential tasks while accumulating and maintaining the knowledge from previous tasks without forgetting. 

Several methods have been proposed to address the CL paradigm with a focus on alleviating the catastrophic forgetting. These methods generally follow three strategies: (1) rehearsal-based methods \cite{shin2017continual,mocanu2016online} maintain the performance of previous tasks by replaying their data during learning new tasks, either the real data or generated one from generative models, (2) regularization-based methods \cite{kirkpatrick2017overcoming,zenke2017continual} aim at using a fixed model capacity and preserving the significant parameters for previous tasks by constraining their change, and (3) architectural-based methods \cite{rusu2016progressive,yoon2018lifelong} dynamically expand the network capacity to reduce the interference between the new tasks and the previously learned ones. Some other methods combine the rehearsal and regularization strategies \cite{pomponi2020efficient,rebuffi2017icarl}. Rehearsal strategies tend to perform well but are not suitable to the situations where one can not access the data from previous tasks (e.g. due to data rights) or where there is computational or storage constraints hinder retaining the data from all tasks (e.g. resource-limited devices). Architectural strategies also achieve a good performance in the CL paradigm at the expense of increasing the model capacity. Regularization strategies utilize a fixed capacity to learn all tasks. However, these methods suffer from significant performance degradation when applied in the class incremental learning (IL) scenario as argued by \cite{kemker2018measuring,Hsu18_EvalCL,Farquhar2019a,VanDeVen2018a}. Following the formulation from \cite{Hsu18_EvalCL,VanDeVen2018a}, in the class IL scenario, the task identity is not available during inference and a unified classifier with a shared output layer (single-headed) is used for all classes. On the other hand, most of the current CL methods assume the availability of the task identity during inference and the model has a separate output layer for each task (multi-headed), a scenario named by \cite{Hsu18_EvalCL,VanDeVen2018a} as task incremental learning.  
Class IL scenario is more challenging; however, class incremental capabilities are crucial for many applications. For example, object recognition systems based on DNNs should be scalable to classify new classes while maintaining the performance of the old classes. Besides, it is more realistic to have all classes sharing the same single-headed output layer without the knowledge of the task identity after deployment.
\begin{figure}[ht]
\vskip 0.2in
\begin{center}
\centerline{\includegraphics[width=0.9\columnwidth]{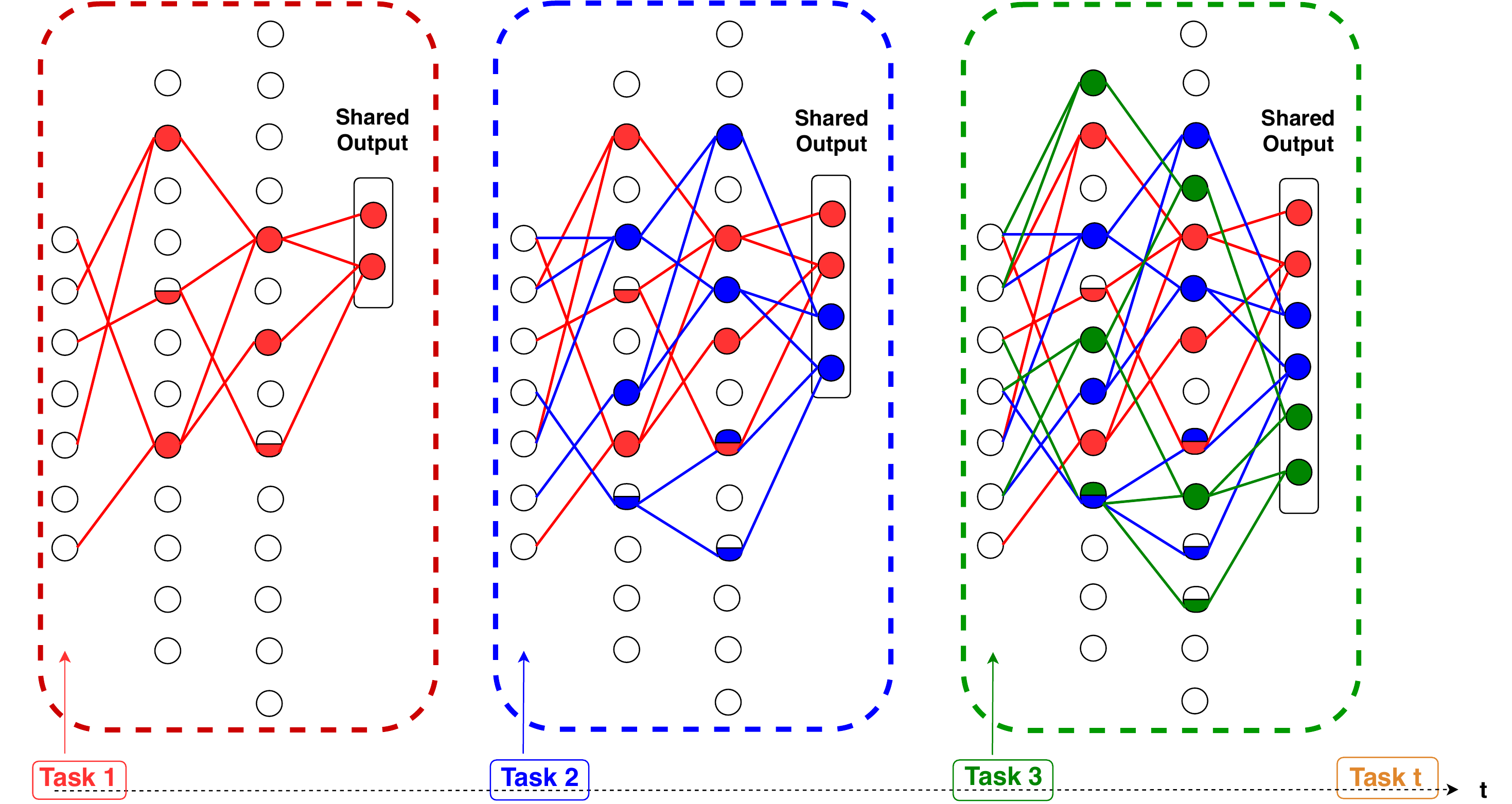}}
\caption{An overview of SpaceNet method for learning a sequence of tasks. All tasks have the same shared output layer. The figure demonstrates the states of the network after learning each of the first three tasks in the sequence. When the model faces a new task $t$, sparse connections are allocated for that task and compacted throughout the sparse adaptive training in the most important neurons, making free space for learning more tasks. The fully filled circles represent the neurons that are most important and become specific for task $t$, where partially filled ones are less important and could be shared by other tasks. Multiple colored circles represent the neurons that are used by multiple tasks. After learning task $t$, the corresponding weights are kept fixed.}
\label{SpaceNet_overview}
\end{center}
\vskip -0.2in
\end{figure}

In this paper, we propose a new architectural-based method for CL paradigm, which we name as SpaceNet. We address the scenario that is not largely explored: class IL in which the model has a single-headed output layer and the task identity is not accessible during inference. We also assume that the data from previous tasks is not available during learning new tasks. Different from previous architectural-based methods, SpaceNet utilizes effectively the fixed capacity of a model instead of expanding the network. The proposed method is based on the adaptive training of sparse neural networks from scratch, a concept introduced by us in \cite{mocanu2018scalable}. The motivation for using sparse neural networks is not only to free space in the model for future tasks but also to produce sparse representations (semi-distributed representations) throughout the adaptive sparse training which reduces the interference between the tasks. An overview of SpaceNet is illustrated in Figure \ref{SpaceNet_overview}. During learning each task, its sparse connections are evolved in a way that compresses them in a compact number of neurons and gradually produces sparse representations in the hidden layers throughout the training. After convergence, some neurons are reserved to be specific for that task while other neurons can be shared with other tasks based on their importance towards the task. This allows future tasks to use the previously learned knowledge during their learning while reducing the interference between the tasks. The adaptive sparse training is based on the readily available information during the standard training, no extra computational or memory overhead is needed to learn new tasks or remember the previous ones.
Our main contributions in this research are:
\begin{itemize}
\item We propose a new method named SpaceNet for continual learning, addressing the more challenging scenario, class incremental learning. SpaceNet utilizes the fixed capacity of the model by compressing the sparse connections of each task in a compact number of neurons throughout the adaptive sparse training. The adaptive training results in sparse representations that reduce the interference between the tasks. 

\item We address more desiderata for continual learning besides alleviating the catastrophic forgetting problem such as memory constraints, computational costs, a fixed model capacity, inaccessibility of previous tasks data, and non-availability of task identity during inference.

\item We achieve a better performance, in terms of robustness to
catastrophic forgetting, than the state-of-the-art regularization and architectural methods using a fixed model capacity, outperforming the regularization methods by a big margin.
\end{itemize}

\section{Related Work}
The interest in CL in recent years has led to a growing number of methods by the research community. The most common methods can be categorized into three main strategies: regularization strategy, rehearsal strategy, and architectural strategy.

\textbf{Regularization methods} aim to protect the old tasks by adding regularization terms in the loss function that constrain the change to neural network weights.
Multiple approaches have been proposed such as: Elastic Weight Consolidation (EWC) \cite{kirkpatrick2017overcoming}, Synaptic Intelligence (SI) \cite{zenke2017continual}, and Memory Aware Synapses (MAS) \cite{aljundi2018memory}. Each of these methods proposed an estimation of the importance of each weight with respect to the trained task. During the training of a new task, any change to the important weights of the old tasks is penalized. 
Learning Without Forgetting (LWF) \cite{li2017learning} is another regularization method that limits the change of model accuracy on the old tasks by using a distillation loss \cite{hinton2014distilling}. The current task data is used to compute the response of the model on old tasks. During learning new tasks, this response is used as a regularization term to keep the old tasks stable. Despite that regularization methods are suitable for the situations where one can not access the data from previous tasks, their performance degrade much in class incremental learning scenario \cite{kemker2018measuring,Hsu18_EvalCL,Farquhar2019a,VanDeVen2018a}. 

\textbf{Rehearsal methods} replay the old tasks data along with the current task data to mitigate the catastrophic forgetting of the old tasks. Deep Generative Replay (DGR) \cite{shin2017continual} trains a generative model on the data distribution instead of storing the original data from previous tasks. Similar work has been done by Mocanu et al. \cite{mocanu2016online}. Other methods combine the rehearsal and regularization strategies such as iCaRL \cite{rebuffi2017icarl}. The authors use distillation loss along with an examplar set to impose output stability of old tasks.  
The main drawbacks of rehearsal methods are the memory overhead of storing old data or a model for generating them, the computational overhead of retraining the data from all previous tasks, and the unavailability of the previous data in some cases. 

\textbf{Architectural methods} modify the model architecture in different ways to make space for new information while keeping the old one. PathNet \cite{fernando2017pathnet} uses a genetic algorithm to find which parts of the
network can be reused for learning new tasks. During the learning of new tasks, the weights of the old tasks are kept frozen. The approach has high computational complexity. CLNP \cite{golkar2019continual} uses a simpler way to find the parts that can be reused in the network by calculating the average activity of each neuron. The least active neurons are reassigned for learning new tasks. Progressive Neural Network (PNN) \cite{rusu2016progressive} is a combination of network expansion and parameter freezing. Catastrophic forgetting is prevented by instantiating a new neural network for each task, while keeping previously learned networks frozen. New networks can take advantage of previous layers learning through the inter-network connections. In this method, the number of model parameters keeps increasing over time. Copy-Weights with Reinit (CWR) \cite{lomonaco2017core50} is counterpart for PNN. The authors proposed an approach that has a fixed model size but has limited applicability and performance. They used fixed shared parameters between the tasks while the output layer is extended when the model faces a new task. Dynamic Expandable Network (DEN) \cite{yoon2018lifelong} keeps the network sparse via weight regularization. Part of the weights of the previous tasks is jointly used with the new task weights to learn the new task. This part is chosen regardless of the importance of it to the old task. If the performance of the old tasks degrades much, they try to restore it by node duplication. Recent methods have been proposed based on sparse neural networks \cite{mallya2018piggyback,mallya2018packnet}. PackNet \cite{mallya2018packnet} prunes the unimportant weights after learning each task and retrains the network to free some connections for later tasks. A mask is saved for each task to specify the connections that will be used during the prediction time. While in the Piggyback method \cite{mallya2018piggyback}, instead of learning the network weights, a mask is learned for each task to select some weights from a pre-trained dense network. These methods require the task identity during the inference to activate the corresponding mask to a test input. Our method is different from these ones in many aspects: (1) we address the class incremental learning scenario where the task identity is unknown during inference, (2) we aim to avoid the computational overhead of iterative pruning and fine-tuning the network after learning each task, and (3) we propose to introduce the sparsity in the representations on the top of the topological sparsity. 

\begin{table}
\caption{Comparison between different CL methods on desired characteristics for CL.}\label{algoComparsion}
\centering
\resizebox{\columnwidth}{!}{%
\begin{tabular}{|l|c|c|c|c|c|c|}  
\hline
Strategy&Method & Fixed Model Capacity & Memory Efficiency & Fast Training & Old Data Inaccessibility & Old Tasks Performance\\
\hline
\multirow{3}{*}{Regularization}&EWC & $\surd$ & $\surd$&$\surd$ & $\surd$ & $\times$\\
&SI & $\surd$ &$\surd$&$\surd$ & $\surd$ & $\times$  \\
&LWF& $\surd$ &$\surd$&$\surd$ & $\surd$ & $\times$  \\
\hline
\multirow{2}{*}{Rehearsal}&iCaRL & $\surd$ & $\times$ & $\surd$ & $\times$ & $\surd$ \\
&DGR   & $\surd$ &$\times$&$\times$ & $\surd$ & $\surd$  \\
\hline
\multirow{4}{*}{Architectural}&PNN & $\times$&$\surd$&$\surd$ & $\surd$ & $\surd$ \\
&PackNet & $\surd$& $\times$& $\times$ & $\surd$ & $\surd$\\
&DEN & $\times$& $\times$& $\times$& $\surd$ &$\surd$   \\
&SpaceNet (Our)   & $\surd$ & $\surd$& $\surd$& $\surd$& $\surd$  \\
\hline
\end{tabular}
}
\end{table}

Most of these works use a certain strategy to address the catastrophic forgetting in the CL paradigm. However, there are more desired characteristics for CL as argued by \cite{schwarz2018progress,Farquhar2019a}. Table \ref{algoComparsion} summarizes a comparison between different algorithms from CL desiderata aspects. The CL algorithm should be constrained in terms of computational and memory overhead. The model size should kept fixed and additional unnecessary neural resources should not be allocated for new tasks. New tasks should be added without adding high computational complexity or retraining the model. The CL problem should be solved without the need for additional memory to save the old data or specific mask to each task.  Lastly, the algorithm should not assume the availability of old data. 

\section{Problem Formulation} 
\label{problemformulation_subsection}
A continual learning problem consists of a sequence of tasks \{$1, 2,...,t,...,T$\}; where $T$ is the total number of tasks. Each task $t$ has its own dataset $D^{t}$. The neural network model faces tasks one by one. The capacity of the model should be utilized to learn the sequence of the tasks without forgetting any of them. All samples from the current task are observed before switching to the next task. The data across the tasks is not assumed to be identically and independently distributed (iid). To handle the situations when one cannot access the data from previous tasks, we assume that once the training of the current task ends, its data becomes not available.

In this work, we address the class incremental learning scenario for CL. In this setting, all tasks share a single-headed output layer. The task identity is not available at deployment time. At any point in time, the network model should classify the input to one of the classes learned so far regardless of the task identity. 

\section{SpaceNet Approach for Continual Learning}
In this section, we present our proposed method, SpaceNet, for deep neural networks to learn in the continual learning paradigm. 

The main objectives of our approach are: (1) utilizing the model capacity efficiently by learning each task in a compact space in the model to leave a room for future tasks, (2) learning sparse representations to reduce the interference between the tasks, and (3) avoiding adding high computational and memory overhead for learning new tasks. In \cite{mocanu2016topological}, we have introduced the idea of training sparse neural networks from scratch for single task unsupervised learning. Lately, this concept has started to be known as sparse training. In recent years, sparse training proved its success in achieving the same performance with dense neural networks for single task standard supervised/unsupervised learning, while having much faster training speed and much lower memory requirements \cite{mocanu2018scalable,bellec2018deep,dettmers2019sparse,evci2019rigging,junjie2019dynamic,mostafa2019parameter}. In these latter works, sparse neural networks are trained from scratch and the sparse network structure is dynamically changed throughout the training. Works from \cite{evci2019rigging,mostafa2019parameter} also show that the sparse training achieves better performance than iteratively pruning a pre-trained dense model and static sparse neural networks. Moreover, Liu et al. \cite{liu2020topological} demonstrated that there is a plenitude of sparse sub-networks with very different topologies that achieve the same performance.

Taking inspiration from these successes and observations, as none of the above discussed sparse training methods are suitable for direct use in continual learning, we propose an adaptive sparse training method for the continual learning paradigm. In particular, in this work, we adaptively train sparse neural networks from scratch to learn each task with a low number of parameters (sparse connections) and gradually develop sparse representations throughout the training instead of having fully distributed representations over all the hidden neurons. Figure \ref{SpaceNet_overview} illustrates an overview of SpaceNet. When the model faces a new task, new sparse connections are randomly allocated between a selected number of neurons in each layer. The learning of this task is then performed using our proposed adaptive sparse training. At the end of the training, the initial distribution of the connections is changed, more connections are grouped in the important neurons for that task. The most important neurons from the initially selected ones are reserved to be specific to this task, while the other neurons are shared between the tasks. The details of our proposed approach are illustrated in Algorithm \ref{spaceNet_alg}. Learning each task in the continual learning sequence by SpaceNet can be divided into 3 main steps: (1) Connections allocation, (2) Task training, (3) Neurons reservation.   
\begin{algorithm}[H]
\small
\caption{SpaceNet for Continual Learning}
\label{spaceNet_alg}
\begin{algorithmic}[1]
\State \textbf{Require:} loss function $\mathcal{L}$ , training dataset for each task in the sequence $\mathcal{D}^{t}$
\State \textbf{Require:} sparsity level $\epsilon$, rewiring fraction $r$
\State \textbf{Require:} number of selected neurons $sel^{t}_{l}$, number of specific neurons $spec^{t}_{l}$

\ForEach{layer $l$}
\State $\textbf{h}^{free}_{l} \gets \textbf{h}_{l}$ \algorithmiccomment{Initialize free neurons with all neurons in $l$ }
\State $\textbf{h}^{spec}_{l} \gets \emptyset$
\State $W_{l} \gets \emptyset$
\State $W^{saved}_{L} \gets \emptyset$
\EndForEach
\ForEach{available task t}
\State $\textbf{W}$ $\gets$ \textit{ConnectionsAllocation}($\epsilon,sel_{l}^{t},\textbf{h}^{free}$)\algorithmiccomment{Perform Algorithm \ref{connection_allocation_alg}}
\State $W^{t}$ $\gets$ \textit{TaskTraining}($\textbf{W}$,$D^{t}$,$\mathcal{L}$,$r$) \algorithmiccomment{Perform Algorithm \ref{task_training_alg}}
\State $\textbf{h}^{free}_{l} \gets$ \textit{NeuronsReservation($spec_{l}^{t}$)} \algorithmiccomment{Perform Algorithm \ref{neurons_reservation_alg}}

\State $W^{saved}_{L} \gets W^{saved}_{L} \cup W^{t}_{L}$ \algorithmiccomment{Retain the connections of last layer for task t }
\State $W_{L} \gets W_{L} \setminus W^{t}_{L}$
\EndForEach
\end{algorithmic}
\end{algorithm}

\textbf{Connections allocation.} Suppose that we have a neural network parameterized by $\textbf{W} = \{W_{l}\}^{L}_{l=1}$, where $L$ is the number of layers in the network. Initially, the network has no connections ($\textbf{W} = \emptyset$). A list of free neurons $\textbf{h}^{free}_{l}$ is maintained for each layer. This list contains the neurons that are not specific for a certain task and can be used by other tasks for connections allocation. When the model faces a new task $t$, the shared output layer $\textbf{h}_{L}$ is extended with the number of classess in this task $n^{t}_{c}$. New sparse connections $W^t = \{W^{t}_{l}\}^{L}_{l=1}$ are allocated in each layer for that task. A selected number of neurons $sel^t_{l}$ (which is hyperparameter) is picked from $\textbf{h}^{free}_{l}$ in each layer for allocating the connections of task $t$. The selected neurons for task $t$ in layer $l$ is represented by $\textbf{h}^{sel}_{l}$. Sparse parameters $W^{t}_{l}$ with sparsity level $\epsilon$ are randomly allocated between $\textbf{h}^{sel}_{l-1}$ and $\textbf{h}^{sel}_{l}$. The parameters $W^{t}$ of task $t$ is added to the network parameters $\textbf{W}$. Algorithm \ref{connection_allocation_alg} describes the connections allocation process. 

\begin{algorithm}[H]
\small
\caption{Connections allocation}
\label{connection_allocation_alg}
\begin{algorithmic}[1]
\State \textbf{Require:} number of selected neurons $sel^{t}_{l}$, sparsity level $\epsilon$
\State $\textbf{h}_{L} \gets$  $\textbf{h}_{L} \cup n_{c}^{t}$ \algorithmiccomment{Expand the shared single output layer with new task classes}
\ForEach{layer}
\State $(\textbf{h}^{sel}_{l-1}, \textbf{h}^{sel}_{l})\gets$ randomly select $sel_{l-l}^{t}$ and $sel_{l}^{t}$ neurons from $\textbf{h}_{l-1}^{free}$ and $\textbf{h}_{l}^{free}$ 
\State randomly allocate parameters $W^{t}_{l}$ with sparsity $\epsilon$ between $\textbf{h}^{sel}_{l-1}$ and $\textbf{h}^{sel}_{l}$
\State $W_{l} \gets W_{l} \cup W^{t}_{l} $  
\EndForEach
\end{algorithmic}
\end{algorithm}

\textbf{Task training.} The task is trained using our proposed adaptive sparse training. The training data $D^{t}$ of task $t$ is forwarded through the network parameters $\textbf{W}$. The parameters of the task $W^{t}$ is optimized with the following objective function:
\begin{equation}
\min_{W^{t}} \mathcal{L}(W^{t};D^{t},W^{1:t-1}), 
\label{optimizeTaskEqu}
\end{equation}
where $\mathcal{L}$ is the loss function and $W^{1:t-1}= \textbf{W} \setminus W^{t}$ are the parameters of the previous tasks. The parameters $W^{1:t-1}$ are freezed during learning task $t$. During the training process, the distribution of sparse connections of task $t$ is adaptively changed, ending up with the sparse connections compacted in a fewer number of neurons. Algorithm \ref{task_training_alg} shows the details of the adaptive sparse training algorithm. After each training epoch, a fraction $r$ of the sparse connections $W_{l}^{t}$ in each layer is dynamically changed based on the importance of the connections and neurons in that layer. Their importance is estimated using the information that is already calculated during the training epoch, no additional computation is needed for importance estimation as we will discuss next. The adaptive change in the connections consists of two phases: (1) Drop and (2) Grow. 

\textbf{Drop phase}. A fraction $r$ of the least important weights is removed from each sparse parameter $W^{t}_{l}$. Connection importance is estimated by its contribution to the change in the loss function. The first-order Taylor approximation is used to approximate the change in loss during one training iteration $i$ as follows:
\begin{equation}
    \mathcal{L}(\textbf{W}^{i+1})-\mathcal{L}(\textbf{W}^{i})\approx \sum_{j=0}^{m-1}{\frac{\partial{\mathcal{L}}}{\partial{{W}^{i}_{j}}}}(W_{j}^{i+1}-W_{j}^{i}) = \sum_{j=0}^{m-1}{I_{i,j}},
\end{equation}
where $\mathcal{L}$ is the loss function, $\textbf{W}$ is the sparse parameters of the network, m is the total number of parameters, and $I_{i,j}$ represents the contribution of the parameter $j$ in the loss change during the step $i$, i.e. how much does a small change to the parameter change the loss function \cite{lan2019lca}. The importance $\Omega^{j}_{l}$ of connection $j$ in layer $l$ at any step is cumulative of the magnitude of $I_{i,j}$ from the beginning of the training till this step.

It is calculated as follows: 

\begin{equation}
\Omega^{j}_{l}=\sum_{i=0}^{iter} ||{I_{i,j}}||,
\label{connectionImp}
\end{equation}
where $iter$ is the current training iteration.

\textbf{Grow phase}. The same fraction $r$ of the removed connections are added in each sparse parameter $W^{t}_{l}$. The newly added weights are zero-initialized. The probability of growing a connection between two neurons in layer $l$ is proportional to the importance of these two neurons $G_{l}$. The importance $a_{l}^{(i)}$ of the neuron $i$ in layer $l$ is estimated by the summation of the importance of ingoing connections of that neuron as follows: 
\begin{equation}
a_{l}^{(i)}=\sum_{j=0}^{C_{in}-1} {\Omega^{j}_{l}},
\label{NeuronImportance}
\end{equation}
where $C_{in}$ is the number of ingoing connections of a neuron $i$ in layer $l$. The matrix $G_{l}$ is calculated as follows:

\begin{equation}
G_{l}= \textbf{a}_{l-1} \textbf{a}^{T}_{l} 
\label{growProbability}
\end{equation}
Assuming that the number of growing connections in layer $l$ is $k_{l}$, the top-$k_{l}$ positions which contains the highest values in $G_{l}$ and zero-value in $W_{l}$ are selected for growing the new connections. 

\begin{algorithm}[H]
\small
\caption{Adaptive sparse training}
\label{task_training_alg}
\begin{algorithmic}[1]
\State \textbf{Require:} loss function $\mathcal{L}$ , training dataset $\mathcal{D}^{t}$, rewiring fraction $r$
\ForEach{training epoch}
\State perform standard forward pass through the network parameters $\textbf{W}$
\State update parameters $W^{t}$ using Equation \ref{optimizeTaskEqu} 
\ForEach{sparse parameter $W^{t}_{l}$} 
\State $\widetilde{W^{t}_{l}}$ $\gets$ sort $W^{t}_{l}$ based on the importance $\Omega_{l}$ in Equation \ref{connectionImp} 
\State ($W^{t}_{l},k_{l}) \gets$ drop ($\widetilde{W^{t}_{l}}$,$r$) \algorithmiccomment{Remove the weights with smallest importance}
\State compute $\textbf{a}_{l-1}$ and $\textbf{a}_{l}$ from Equation \ref{NeuronImportance} \algorithmiccomment{Neurons importance for task t}
\State $G_{l} \gets$ $\textbf{a}_{l-1} \textbf{a}^{T}_{l}$ 
\State $\widetilde{G_{l}} \gets$ sortDescending($G_{l}$)
\State \textit{Gpos} $\gets$ select top-$k_{l}$ positions in $\widetilde{G_{l}}$ where $W_{l}$ equals zero 
\State $W^{t}_{l} \gets$ grow($W^{t}_{l}$,G\textit{pos}) \algorithmiccomment{Grow $k_{l}$ zero-initialized weights in \textit{Gpos}} 
\EndForEach
\EndForEach
\end{algorithmic}
\end{algorithm}

For convolutional neural networks, the drop and grow phases are performed in a coarse manner to impose structure sparsity instead of irregular sparsity. In particular, in the drop phase, we consider coarse removal for the whole kernel instead of removing scalar weights. The kernel importance is calculated by the summation over the importance of its $k\times k$ elements calculated by Equation \ref{connectionImp}. Similarly, in the grow phase, the whole connections of a kernel are added instead of adding single weights. Analogous to multilayer perceptron networks, the probability of adding a kernel between two feature maps is proportional to their importance. The importance of the feature map is calculated by the summation of the importance of its connected kernels. 

\textbf{Neurons reservation}. After learning the task, a fraction of the neurons from $\textbf{h}^{sel}_{l}$ in each layer is reserved for this task and removed from the list of free neurons $\textbf{h}^{free}_{l}$. The choice of these neurons is based on their importance to the current task calculated by equation \ref{NeuronImportance}. These neurons become specific to the current task which means that no more connections from other tasks will go in these neurons. The other neurons in $\textbf{h}^{sel}_{l}$ are still exist in the free list $\textbf{h}^{free}_{l}$ and could be shared by future tasks. Algorithm \ref{neurons_reservation_alg} describes the details of neurons reservation process. 

\begin{algorithm}[H]
\small
\caption{Neurons reservation}
\label{neurons_reservation_alg}
\begin{algorithmic}[1]
\State \textbf{Require:} number of specific neurons $spec^{t}_{l}$
\ForEach{layer $l$}
\State compute the neuron importance $a_{l}$ for task t using Equation \ref{NeuronImportance}
\State $\widetilde{\textbf{a}_{l}} \gets $ sortDescending($\textbf{a}_{l}$)
\State  $\textbf{h}_{l}^{t_{spec}}$ $ \gets$ top-$spec_{l}^{t}$ from $\widetilde{a_{l}}$
\State $\textbf{h}^{spec}_{l} \gets \textbf{h}^{spec}_{l} \cup \textbf{h}_{l}^{t_{spec}}$
\State $\textbf{h}^{free}_{l} \gets \textbf{h}^{free}_{l} \setminus \textbf{h}_{l}^{t_{spec}}$
\EndForEach
\end{algorithmic}
\end{algorithm}

After learning each task, its sparse connections in the last layer (classifier) are removed from the network and retained aside in $W_{L}^{saved}$. Removing the classifiers ($W^{1:t-1}_{L}$) of the old tasks during learning the new one contributes to alleviating the catastrophic forgetting problem. If they are all kept, the weights of the new task will try to get higher values than the weights of the old tasks to be able to learn which results in a bias towards the last learned task during inference.
At deployment time, the output layer connections $W_{L}^{saved}$ for all learned tasks so far are returned to the network weights $W_{L}$. All tasks share the same single-headed output layer. 

\paragraph{Link to Hebbian Learning} The way we evolve the sparse neural network during the training of each task has a connection to Hebbian learning. Hebbian learning \cite{hebb1949organization} is considered as a plausible theory for biological learning methods. It is an attempt to explain the adaptation of brain neurons during the learning process. The learning is performed in a local manner. The weight update is not based on the global information of the loss. The theory is usually summarized as \enquote{cells that fire together wire together}. It means that if a neuron participates in the activation of another neuron, the synaptic connection between these two neurons should be strengthened. Analogous to Hebb's rule, we consider changing the structure of the sparse connections in a way that increases the number of connections between strong neurons. 
\section{Experiments}
  We compare SpaceNet with well-known approaches from different CL strategies. The goals of this experimental study are: (1) evaluating SpaceNet ability in maintaining the performance of previous tasks in the class IL scenario using two typical DNN models  (i.e. multilayer perceptron and convolutional neural networks), (2) analyzing the effectiveness of our proposed adaptive sparse training in the model performance, and (3) comparing between different CL methods in terms of performance and other requirements of CL such as model size and using extra memory. We evaluated our proposed method on four well-known benchmarks for continual learning: split MNIST \cite{lecun1998mnist,zenke2017continual}, split Fashion-MNIST \cite{xiao2017fashion,Farquhar2019a}, CIFAR-10/100 \cite{krizhevsky2009learning,zenke2017continual}, and iCIFAR-100 \cite{krizhevsky2009learning,rebuffi2017icarl}. We used two metrics for evaluating our proposed CL method. The first one, ACC, is the average classification accuracy across all tasks. The second one is the backward transfer metric \cite{lopez2017gradient}, BWT, which measures the influence of learning new tasks on the performance of previous tasks. Larger negative value for BWT indicates catastrophic forgetting. Formally, the ACC and BWT are calculated as follows:
\begin{equation}
\label{ACC_BWT}
\begin{split}
   ACC = \frac{1}{T} \sum_{i=1}^{T} R_{T,i}, \\
   BWT = \frac{1}{T-1} \sum_{i=1}^{T-1} R_{T,i} - R_{i,i}, 
\end{split}
\end{equation}
where $R_{j,i}$ is the accuracy on task ${i}$ after learning
the $j$-th task in the sequence, and $T$ is the total number of tasks.
\subsection{Split MNIST}
Split MNIST is first introduced by Zenke et al. \cite{zenke2017continual}. It consists of five tasks. Each task is to distinguish between two consecutive MNIST-digits. This dataset becomes a commonly used benchmark for evaluating continual learning approaches. Most authors use this benchmark in the multi-headed form where the prediction is limited to two classes only, determined by the task identity during the inference. While for our settings, the input image has to be classified into one of the ten MNIST-digits from 0 to 9 (single-headed layer). 

\subsubsection{Experimental Setup}
The standard training/test-split for MNIST was used resulting in 60,000 training images and 10,000 test images. For a fair comparison, our model has the same architecture used by Van et al. \cite{VanDeVen2018a}. The architecture is a feed-forward network with 2 hidden layers. Each layer has 400 neurons with ReLU activation. We use this fixed capacity to learn all tasks. 10\% of the network weights are used for all tasks (2\% for each task). The rewiring fraction $r$ equals to 0.2. Each task is trained for 4 epochs. We use a batch size of 128. The network is trained using stochastic gradient descent with a learning rate 0.01. The selected number of neurons $sel^{t}_{l}$ in each hidden layer to allocate the connections for a new task is 80. The number of neurons that are reserved to be specific for each task $spec^{t}_{l}$ is 40. The hyperparameters are selected using random search. The experiment is repeated 10 times with different random seeds.

\subsubsection{Results}
Table \ref{acc-comparsion-SplitMNIST} shows the average accuracy (ACC) and the backward transfer (BWT) of different well-known approaches. As illustrated in the table, regularization methods fail to maintain the performance of the previously learned tasks in the class IL scenario. They have the lowest BWT performance. The experiment shows that SpaceNet is capable of achieving very good performance. It manages to keep the performance of previously learned tasks, causing a much lower negative backward transfer and outperforming the regularization methods in terms of average accuracy by a big gap around 51.6\%.
\begin{table}
\caption{ACC and BWT on split MNIST using different approaches. Results for regularization and rehearsal methods are adopted from \cite{VanDeVen2018a,Hsu18_EvalCL} except \enquote{SpaceNet-Rehearsal}.}
\label{acc-comparsion-SplitMNIST}
\centering
\resizebox{\columnwidth}{!}{
\begin{tabular}{|l|c|c|c|c|c|c|}  
\hline
Strategy &Method& ACC (\%) & BWT (\%) & Extra memory & Old task data& Model expansion\\
\hline
\multirow{4}{*}{Regularization}&EWC & \textbf{20.01} $\pm$ \footnotesize{0.06} & \textbf{-99.64} $\pm$ \footnotesize{0.01} &\multirow{4}{*}{No} & \multirow{4}{*}{No}& \multirow{4}{*}{No}\\
&SI &  19.99 $\pm$ \footnotesize{0.06}& -99.62 $\pm$ \footnotesize{0.11}& & &\\
&MAS & 19.52 $\pm$ \footnotesize{0.29}& -99.73 $\pm$ \footnotesize{0.06} & & &\\
\hline
\multirow{3}{*}{Rehearsal}&DGR & 90.79 $\pm$ \footnotesize{0.41}& -9.89 $\pm$ \footnotesize{1.02}  & \multirow{3}{*}{Yes} & \multirow{3}{*}{Yes} & \multirow{3}{*}{No}\\
&iCaRL & 94.57 $\pm$ \footnotesize{0.11} & -3.27 $\pm$ \footnotesize{0.14}& & &\\
&SpaceNet-Rehearsal & \textbf{95.08} $\pm$ \footnotesize{0.15}  & \textbf{-3.13} $\pm$ \footnotesize{0.06} & & &\\
\hline
\multirow{3}{*}{Architectural}& DEN & 56.95 $\pm$ \footnotesize{0.02}& -21.71 $\pm$ \footnotesize{1.29} &Yes & No & Yes\\
&Static-SparseNN & 61.25 $\pm$ \footnotesize{2.30} & -29.32 $\pm$ \footnotesize{2.80} &\multirow{2}{*}{\textbf{No}} & \multirow{2}{*}{\textbf{No}} & \multirow{2}{*}{\textbf{No}}\\
& SpaceNet & \textbf{75.53} $\pm$ \footnotesize{1.82}& \textbf{-15.99} $\pm$ \footnotesize{1.83} & & &\\
\hline
\end{tabular}
}
\end{table}
We compare our method also to the DEN algorithm which is the most related one to our work, both being architectural strategies. As discussed in the related work section, DEN keeps the connections sparse by sparse-regularization and restores the drift in old tasks performance using node duplication. In the DEN method, the connections are remarked with a timestamp (task identity) and in the inference, the task identity is required to test on the parameters that are trained up to this task identity only. This implicitly means that $T$ different models are obtained using DEN, where $T$ is the total number of tasks. To make the comparison, we adapt the official code provided by the authors to work on the class IL scenario, where there is no access to the task identity during inference. After training all tasks, the test data is evaluated on the model created each timestamp $t$. The class with the highest probability from all models is taken as the final prediction. Besides that DEN has computational overhead for learning a new task when comparing to SpaceNet, it also increases the number of neurons in each layer by around 35 neurons, while SpaceNet still has unused neurons in the originally allocated capacity; 92 and 91 neurons in the first and second hidden layers respectively. As shown in the table, SpaceNet obtains the best performance and the lowest forgetting among the methods from its strategy (category), reaching an accuracy of about 75.53\%, with 18.5\% better than the DEN algorithm.

Rehearsal methods succeeded in maintaining their performance. Replaying the data from previous tasks during learning a new task mitigates the problem of catastrophic forgetting, hence these methods have the highest BWT performance. However, retraining old tasks data has a cost of requiring additional memory for storing the data or the generative model in case of generative replay methods. Making rehearsal methods resource-efficient is still an open research problem. The results of SpaceNet in terms of both ACC and BWT are considered very satisfactory and promising compared to rehearsal methods given that we do not use any of the old tasks data and the number of connections is much smaller i.e. SpaceNet has 28 times fewer connections than DGR. 

Please note that it is easy to combine SpaceNet with rehearsal strategies. We perform an experiment in which the old tasks data are repeated during learning new tasks, while keeping the connections of the old tasks fixed. We refer to this experiment as \enquote{SpaceNet-Rehearsal}. Replaying the old data helps to find weights for the new task that do not degrade the performance of the old tasks. As shown in Table \ref{acc-comparsion-SplitMNIST}, \enquote{SpaceNet-Rehearsal} outperforms all the state-of-the-art methods, including the rehearsal ones, while having a much smaller number of connections. However, replaying the data from the previous tasks is outside the purpose of this paper where we try besides maximizing performance to cover the scenarios when one has no access to the old data, minimize memory requirements, and reduce the computational overhead for learning new tasks or remember the previous ones.

\begin{figure}[ht]
\vskip 0.2in
\begin{center}
\centerline{\includegraphics[width=0.6\columnwidth]{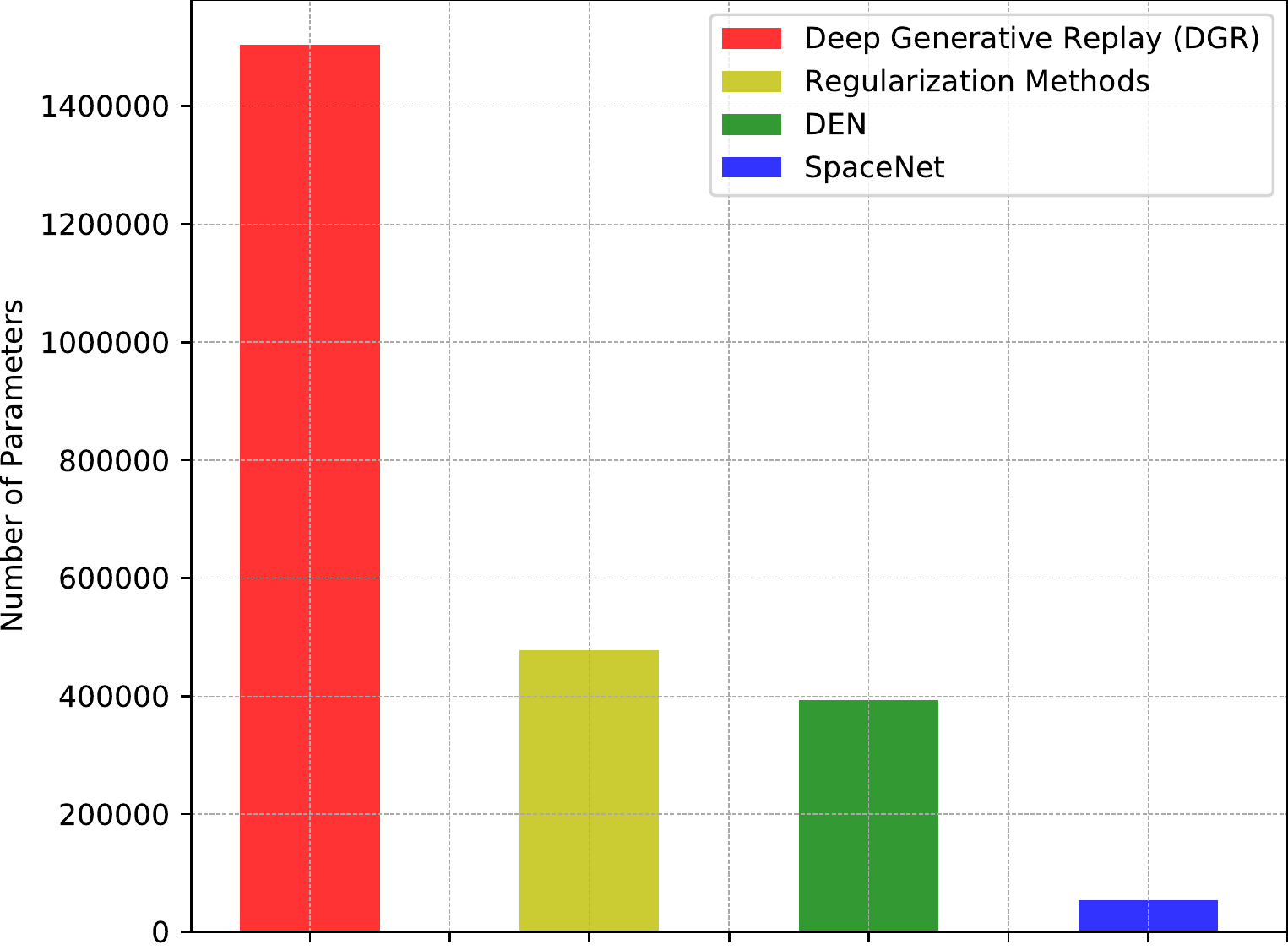}}
\caption{Comparison between SpaceNet and other CL methods on split MNIST in terms of model size.}
\label{SpaceNetvsotherParam}
\end{center}
\vskip -0.2in
\end{figure}

A comparison between different methods in terms of other requirements for CL is also shown in Table \ref{acc-comparsion-SplitMNIST}. Regularization methods satisfy many desiderata of CL while losing the performance. SpaceNet is able to compromise between the performance and other requirements that are not even satisfied by other architectural methods. Moreover, we compare the model size of our approach with the other methods. As illustrated in Figure \ref{SpaceNetvsotherParam}, SpaceNet model with at least one order of magnitude fewer parameters than any of the other method studied.

We further analyze the effect of our proposed adaptive sparse training in performance. We compare our approach with another baseline, referred as \enquote{Static-SparseNN}. In this baseline, we run our proposed approach for CL but with static sparse connections and train the model with the standard training process. As shown in Table \ref{acc-comparsion-SplitMNIST}, the adaptive sparse training increases the performance of the model by a good margin. The average accuracy for all tasks is increased by$~14.28\%$, while the backward transfer performance is increased by 13.3\%.

\subsection{Split Fashion-MNIST}
An additional experiment for validating our approach is performed on the Fashion-MNIST dataset \cite{xiao2017fashion}. This dataset is more complex than MNIST. The images show individual articles of clothing. The authors argued that it is considered as a drop-in replacement for MNIST. However, it has the same sample size and structure of training and test sets as MNIST. This dataset is used by Farquhar and Gal \cite{Farquhar2019a} to evaluate different CL approaches. They construct split Fashion-MNIST which consists of five tasks. Each task has two consecutive classes of Fashion-MNIST. \subsubsection{Experimental Setup}
The same setting and architecture used for the MNIST dataset are used in this experiment except that each task is trained for 20 epochs. We use the official code from \cite{VanDeVen2018a} to test the performance of their implemented CL approaches on split Fashion-MNIST. We do not change the experimental settings to evaluate the performance of the methods on a more complex dataset using such small neural networks. 
\begin{table}
\caption{ACC and BWT on split Fashion-MNIST using different approaches.}
\label{acc-comparsion-FashionMNIST}
\centering
\resizebox{\columnwidth}{!}{
\begin{tabular}{|l|c|c|c|c|c|c|}  
\hline
Strategy &Method& ACC (\%) & BWT (\%) & Extra memory & Old task data& Model expansion\\
\hline
\multirow{3}{*}{Regularization}&EWC & 19.47 $\pm$ \footnotesize{0.98}  & -99.13 $\pm$ \footnotesize{0.39} & \multirow{3}{*}{No} & \multirow{3}{*}{No} & \multirow{3}{*}{No}\\
&SI &  19.93 $\pm$ \footnotesize{0.01}& -99.08 $\pm$ \footnotesize{0.51} & & &\\
&MAS & \textbf{19.96} $\pm$ \footnotesize{0.01} & 	\textbf{-98.82} $\pm$ \footnotesize{0.10} & & &\\
\hline
\multirow{3}{*}{Rehearsal}&DGR & 73.58 $\pm$ \footnotesize{3.90}& -32.56 $\pm$ \footnotesize{3.74} & \multirow{3}{*}{Yes} & \multirow{3}{*}{Yes} & \multirow{3}{*}{No}  \\
&iCaRL & 80.70 $\pm$ \footnotesize{1.29}&  -10.39 $\pm$ \footnotesize{1.97} & & &\\
&SpaceNet-Rehearsal & \textbf{84.18} $\pm$ \footnotesize{0.24} & \textbf{-3.09} $\pm$ \footnotesize{0.24} & & &\\
\hline
\multirow{3}{*}{Architectural} & DEN & 31.51 $\pm$ \footnotesize{0.04} & -47.94 $\pm$ \footnotesize{1.69} & Yes & No & Yes\\ 
&Static-SparseNN & 56.80 $\pm$ \footnotesize{2.30}& -29.79 $\pm$ \footnotesize{2.22} &\multirow{2}{*}{\textbf{No}} & \multirow{2}{*}{\textbf{No}} & \multirow{2}{*}{\textbf{No}}\\
&SpaceNet & \textbf{64.83} $\pm$ \footnotesize{0.69} & \textbf{-23.98} $\pm$ 1.89 &  &  & \\
\hline
\end{tabular}
}
\end{table}
\subsubsection{Results}
We observe the same findings that regularization methods fail to remember previous tasks. The average accuracy of rehearsal methods on this more difficult dataset starts to deteriorate and the negative backward transfer increases. Replaying the data with the SpaceNet approach achieves the best performance. As shown in Table \ref{acc-comparsion-FashionMNIST}, while the accuracy of DEN degrades much, SpaceNet maintains a stable performance on the tasks reaching ACC of 64.83\% and BWT of -23.98\%. The DEN algorithm expands each hidden layer by 37 neurons, while SpaceNet still have 90 and 93 unused neurons in the first and second hidden layers respectively. The sparse training in SpaceNet increases the ACC and BWT by 8\% and 6\% respectively compared to \enquote{Static-SparseNN}.

\subsection{CIFAR-10/100}
In this experiment, we show that our proposed approach can be applied also to convolutional neural networks (CNNs). We evaluate spaceNet on complex datasets: CIFAR-10 and CIFAR-100 \cite{krizhevsky2009learning}. CIFAR-10 and CIFAR-100 are well-known benchmarks for classification tasks. They contain tiny natural images of size ($32\times32$). CIFAR-10 consists of 10 classes and has 60000 samples (50000 training + 10000 test), with 6000 images per class. While CIFAR-100 contains 100 classes, with 600 images per class (500 train + 100 test). Zenke et al. \cite{zenke2017continual} uses these two datasets to create a benchmark for CL which they referred as CIFAR-10/100. It has 6 tasks. The first task contains the full dataset of CIFAR-10, while each subsequent task contains 10 consecutive classes from CIFAR-100 dataset. Therefore, task 1 has a 10x larger number of samples per class which makes this benchmark challenging as the new tasks have limited data.

\subsubsection{Experimental Setup}
\label{cifar_10_experiment_setup}
For a fair and direct comparison, we follow the same architecture used by Zenke et al. \cite{zenke2017continual} and Maltoni and  Lomonaco \cite{maltoni2019continuous}. The architecture consists of 4 convolutional layers (32-32-64-64 feature maps). The kernel size is $3\times3$. Max pooling layer is added after each 2 convolutional layers. Two sparse feed-forward layers follow the convolutional layers (512-60 neurons), where 60 is the total number of classes from all tasks. We replace the dropout layers with batch normalization \cite{ioffe2015batch}. The model is optimized using stochastic gradient descent with learning rate 0.1. Each task is trained for 20 epochs. 12\% of the network weights is used for each task. Since the number of feature maps in each layer in the used architecture is too small, the number of selected feature maps for each task $sel^{t}_{l}$ equals to the number of feature maps in this layer excluding the specific neurons in that layer. The number of specific feature maps in each hidden layer $spec^{t}_{l}$ is as follows: [2, 2, 5, 6, 30]. The hyperparameters are selected using random search.

\begin{figure}[t]
\vskip 0.2in
\begin{center}
\centerline{\includegraphics[width=0.56\columnwidth]{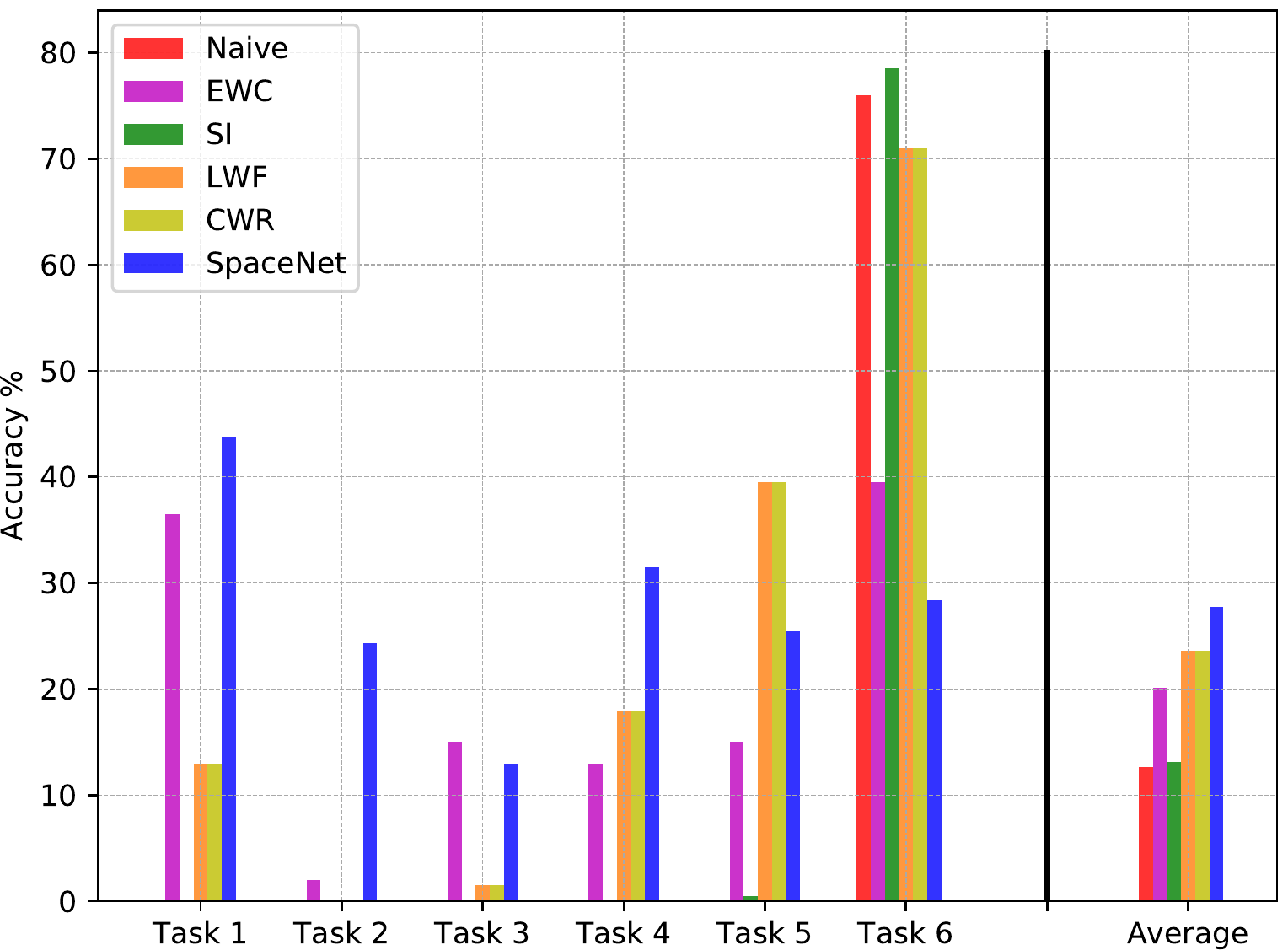}}
\caption{Accuracy on each task of CIFAR-10/100 benchmark for different CL approaches after training the last task. Results for other approaches are adopted from Maltoni and Lomonaco \cite{maltoni2019continuous}. Task 1 is the full dataset of CIFAR10, while task 2 to task 6 are the first 5 tasks from CIFAR100. Each task contains 10 classes. The missing rectangles for some of the methods for some of the tasks means that accuracy for that particular case is 0. The \enquote{Average} x-axis label shows the average accuracies computed overall tasks for each method. SpaceNet managed to utilize the available model capacity efficiently between the tasks, unlike other methods that have high performance on the last task but completely forgetting some other previous tasks.}
\label{Acc_cifar10_100}
\end{center}
\vskip -0.2in
\end{figure}

\subsubsection{Results}
Figure \ref{Acc_cifar10_100} shows the accuracy of different popular CL methods for each task of CIFAR-10/100 after training all tasks. The results of other algorithms are extracted from the work done by Maltoni and Lomonaco \cite{maltoni2019continuous} and re-plotted. \enquote{Naive} algorithm is referred by the authors to the simple finetuning where there is no limitation for forgetting other than early stopping. SI totally fails to remember all old tasks and the model is fitted just on the last learned one. Other algorithms have a good performance on some tasks, while the performance on the other tasks is very low. Despite that the architecture used in this experiment is small, SpaceNet managed to utilize the available space efficiently between the tasks. As the figure shows, SpaceNet outperforms all the other algorithms in terms of average accuracy. In addition, the standard deviation over all tasks accuracy is much (few times) smaller than the standard deviation of any other state-of-the-art method. This means that the model is not biased towards final learned task and the accuracy of the learned tasks is close to each other. This clearly highlights the robustness of SpaceNet and its strong capabilities in remembering old tasks.

This experiment shows that SpaceNet utilizes the small available capacity. Yet, the model capacity could reach its limit after learning a certain number of tasks. In this case, we can allocate more resources (units) to the network to fit more tasks since we have fully utilized the existing ones. To show this case on the same architecture and settings, we increase the number of sparse connections allocated for each task in the second layer to 16.5\% of the layer weights. This leads that the second layer will approximately reach its maximum capacity after learning the first five tasks. When the model faces the last task of the CIFAR-10/100 benchmark, we allocate 8 new features maps in the second and third convolutional layers. The algorithm of SpaceNet continues normally to learn this task. The average accuracy achieved in this experiment equals to 27.89 $\pm$ 0.84. 

\subsection{iCIFAR-100}
In this experiment, we evaluate our proposed method on another CL benchmark with larger number of classes. iCIFAR-100 \cite{rebuffi2017icarl} is a variant of CIFAR-100 \cite{krizhevsky2009learning} which contains 100 classes. This dataset is divided into 5 tasks. Each task has 20 consecutive classes from CIFAR-100. The goal of this experiment is to analyze the behavior of SpaceNet and the regularization methods on larger dataset using two types of CNN architectures: the small CNN network detailed in Section \ref{cifar_10_experiment_setup} (named as \enquote{small CNN}) and a more sophisticated architecture; Wide Residual Networks (WRN) \cite{zagoruyko2016wide}.

\subsubsection{Experimental Setup}
  For the small CNN model, the number of selected feature maps for each task $sel^{t}_{l}$ equals the number of feature maps in this layer excluding the specific feature maps of previous tasks. The number of specific feature maps in each hidden layer $spec^{t}_{l}$ is as follows: [2, 2, 3, 4, 20]. 15\% of the network weights are allocated for each task. For Wide ResNet \cite{zagoruyko2016wide}, we used WRN-28-10, with a depth=28 and widen-factor=10. Since the first group in the residual network has a small number of feature maps, the connections for each task span all the feature maps except the specific ones. $sel^{t}_{l}$ for the other three groups is as follows: [40, 60, 120]. The number of specific feature maps $spec^t_l$ in each group is: [0, 16, 24, 60]. 2\% of the network weights are allocated for each task. The rewiring fraction $r$ equals to 0.3. The two studied models are optimized using stochastic gradient descent with a learning rate 0.1. Each task is trained for 20 epochs. The hyperparameters are selected using random search. Each experiment is repeated 5 times with different random seeds. We use the official code from \cite{Hsu18_EvalCL} to test the performance of their implemented regularization methods using CNNs on this benchmark.
\subsubsection{Results}
Table \ref{acc-comparsion-icifar100} shows the average accuracy (ACC) and the backward transfer (BWT) using the two described architectures. As illustrated in the table, SpaceNet managed to utilize the available capacity of the small CNN architecture achieving higher accuracy than the regularization methods by 13.5\%. SpaceNet is also more robust to forgetting, the BWT is better than the regularization strategy by 19\%. The experiment also shows that allocating a larger network at the beginning of learning the CL sequence does not help the regularization strategy to alleviate the catastrophic forgetting problem. A small increase is gained in the average accuracy due to achieving higher performance in the last task using the larger model, while the forgetting (negative backward transfer) is increased by around 11\%. On the other hand, SpaceNet takes advantage of the more available resources in the larger network (WRN-28-10). The ACC of SpaceNet is increased by 6\% and the forgetting is decreased by 11.5\%.

\begin{table}
\caption{ACC and BWT on the iCIFAR100 benchmark using two different architectures.}
\label{acc-comparsion-icifar100}
\centering
\resizebox{\columnwidth}{!}{
{\begin{tabular}{|l|c|c|c|c|c|}  
\hline
\multicolumn{2}{|c|}{} & \multicolumn{2}{c|}{Small CNN} & \multicolumn{2}{c|}{WRN-28-10}\\
\hline
Strategy &Method& ACC (\%)& BWT (\%) & ACC (\%) &BWT (\%) \\
\hline
\multirow{3}{*}{Regularization}&EWC & 13.65 $\pm$ \footnotesize{0.15}  & 	-64.38 $\pm$ \footnotesize{0.71}  & 16.09 $\pm$ \footnotesize{0.29}& -73.10 $\pm$ \footnotesize{1.11}\\
&SI & 14.45 $\pm$ \footnotesize{0.11} & -66.47 $\pm$ \footnotesize{0.49} & 16.75 $\pm$ \footnotesize{0.30} & 	-77.75 $\pm$ \footnotesize{0.90}  \\
&MAS & 14.51 $\pm$ \footnotesize{0.22} & -66.61 $\pm$ \footnotesize{0.31} & 16.51 $\pm$ \footnotesize{0.20} & -76.82 $\pm$ \footnotesize{0.19}  \\
\hline
Architectural & SpaceNet & 	\textbf{28.11} $\pm$ \footnotesize{0.74} & \textbf{-45.62} $\pm$ \footnotesize{0.88} & \textbf{34.10} $\pm$ \footnotesize{0.92} & \textbf{-34.18} $\pm$ \footnotesize{0.91} \\
\hline
\end{tabular}}
}
\end{table}

\section{Analysis}
In this section, we analyze the representations learned by SpaceNet, the distribution of the sparse connections after the adaptive sparse training, and the relation between the learned distribution of the connections and the importance of the neurons. We performed this analysis on the Split MNIST benchmark. 

\begin{figure}[ht]
 \centering
 \begin{subfigure}[b]{0.45\columnwidth}
     \centering
     \includegraphics[width=\linewidth]{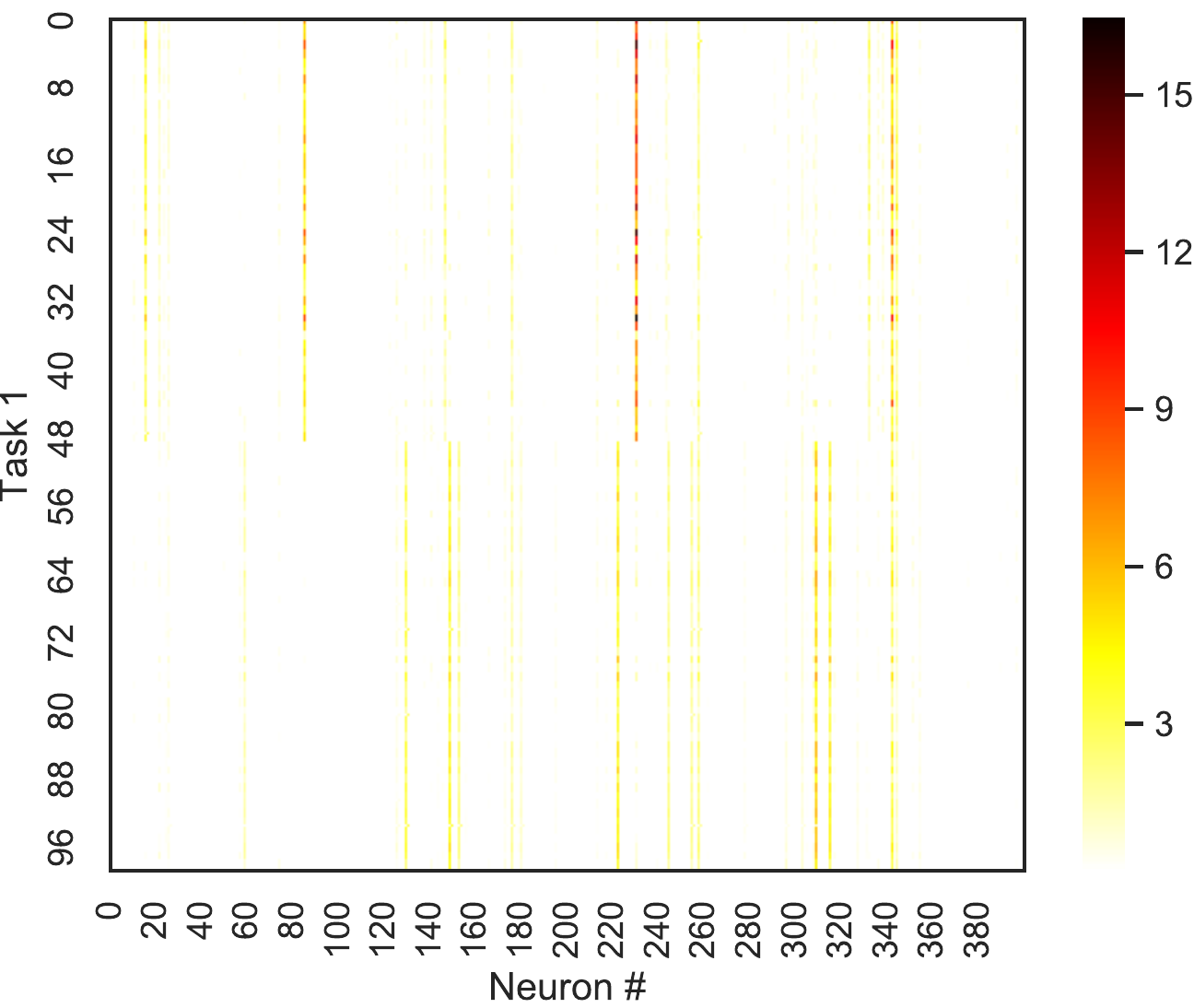}
     \caption{First hidden layer.}
\end{subfigure}
 \begin{subfigure}[b]{0.45\columnwidth}
    \centering
    \includegraphics[width=\linewidth]{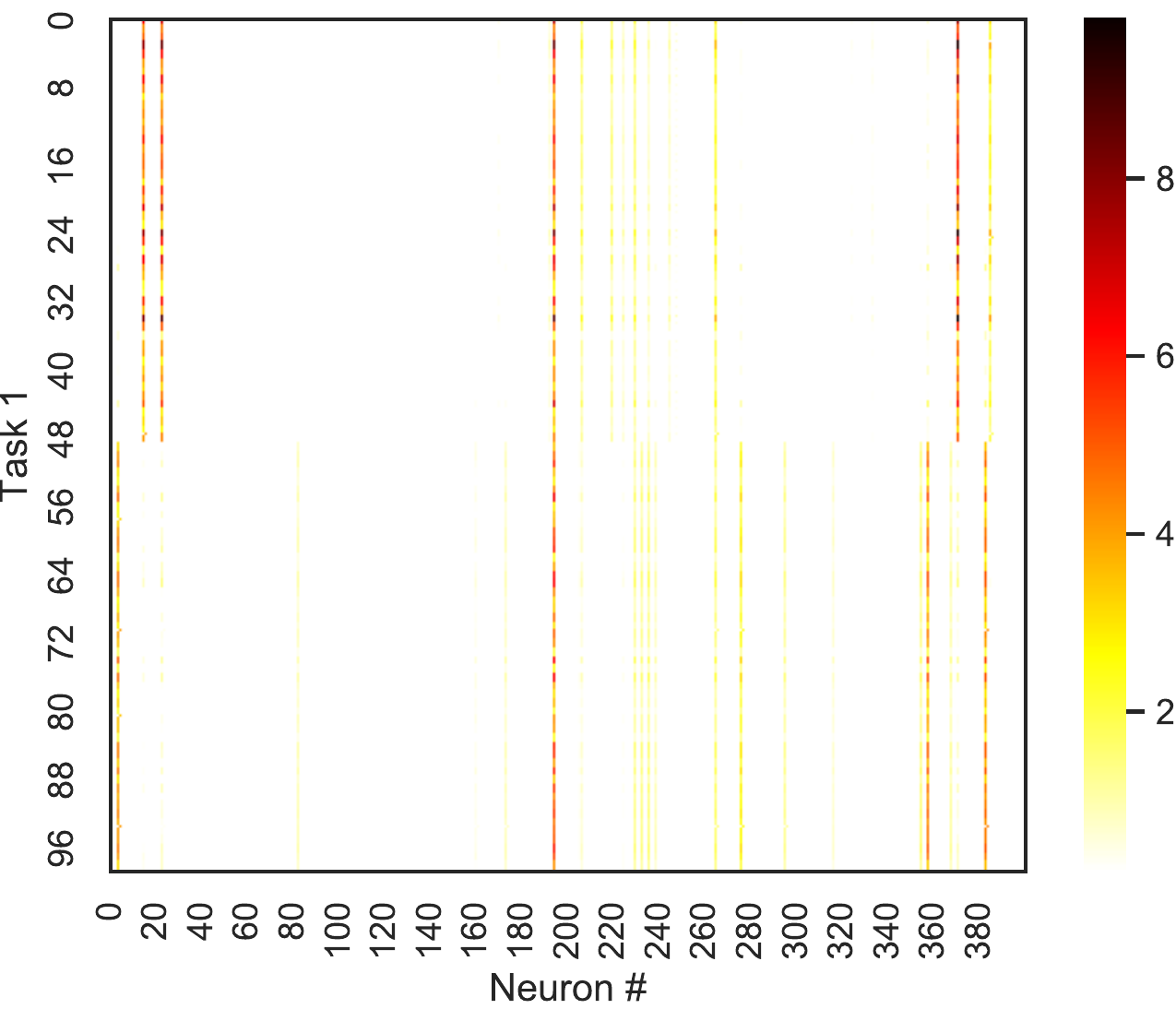}
      \caption{Second hidden layer.}
 \end{subfigure}
\caption{Heatmap of the first and second hidden layers activations after forwarding the test data of task 1 of split MNIST. The y-axis represents the test samples. The first 50 samples belong to class 0 while the other 50 belong to class 1.}
\label{HeatMap}
\end{figure}
       
\begin{figure}[ht]
 \centering
 \begin{subfigure}[b]{0.45\columnwidth}
     \centering
     \includegraphics[width=\linewidth]{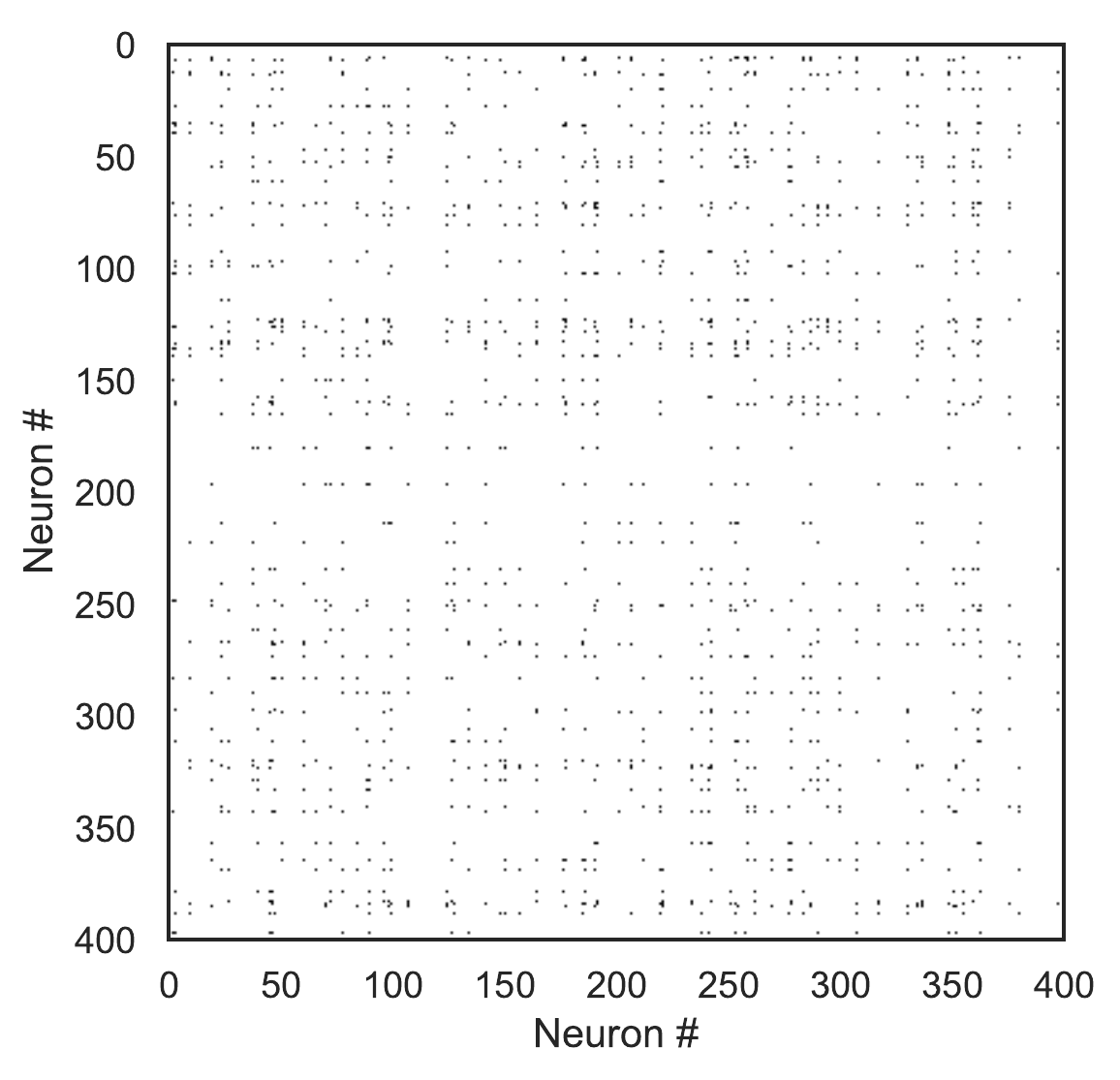}
     \caption{Initial connections.}
     \label{initial_W}
\end{subfigure}
 \begin{subfigure}[b]{0.45\columnwidth}
    \centering
    \includegraphics[width=\linewidth]{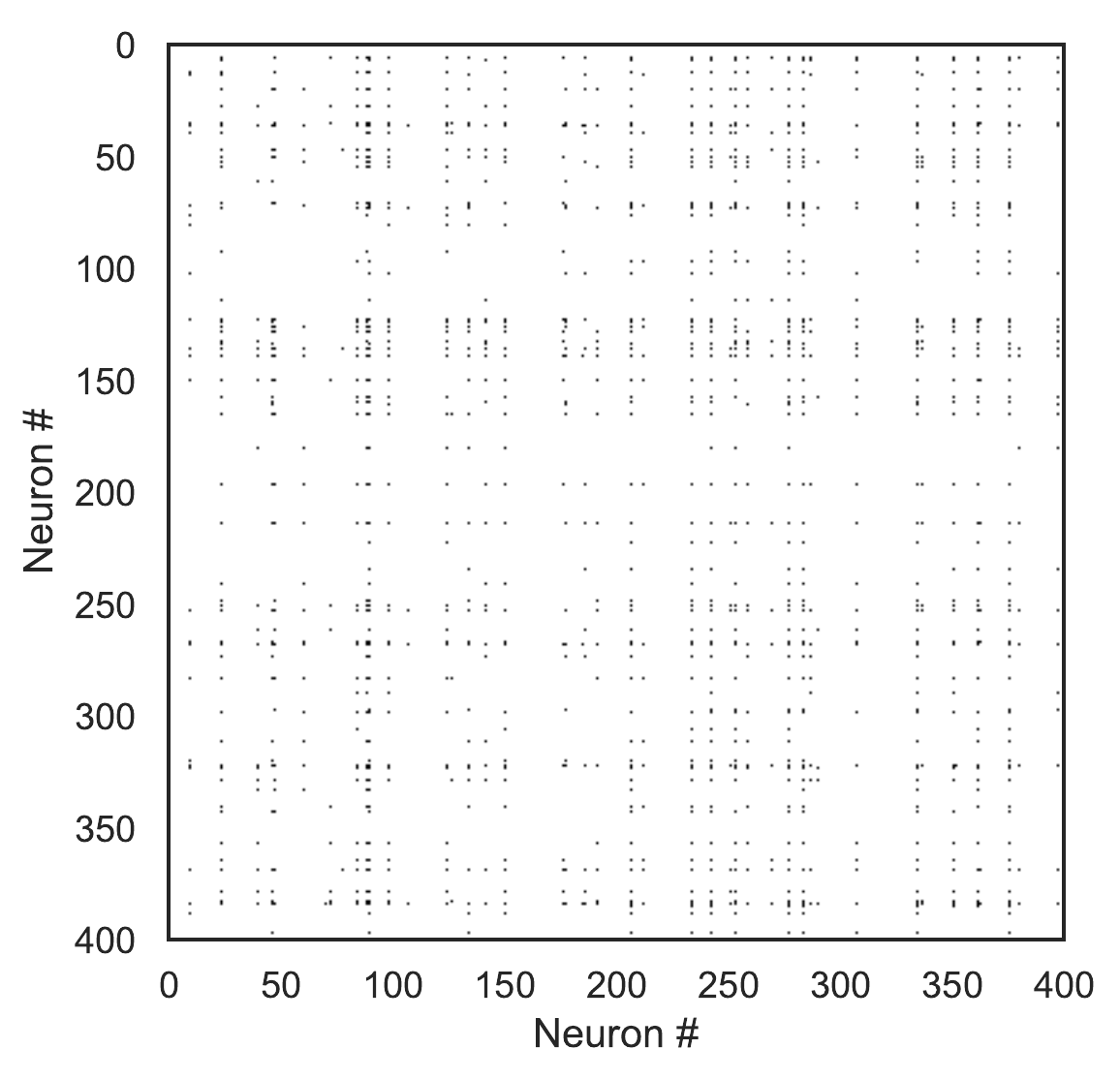}
      \caption{Learned connections.}
     \label{W_after_train}
 \end{subfigure}
 \caption{Connections distribution between two layers for one task of the Split MNIST benchmark. Figure (a) shows the initial random distribution of the connections on the selected neurons. Figure (b) shows the connections after the adaptive sparse training. The connections are compacted in some of the neurons. }
 \label{w_before_after}
\end{figure}

First, we analyze the representations learned by SpaceNet. We visualize the activations of the two hidden layers of the multilayer perception network used for Split MNIST. After learning the first task of Split MNIST, we analyze the representations of random test samples from this task. Figure \ref{HeatMap} shows the representations of 50 random samples from the test set of class 0 and another 50 samples from the test set of class 1. The figure illustrates that the representations learned by SpaceNet are highly sparse. A small percentage of activations is used to represent an input. This reveals that the designed topological sparsity of SpaceNet not only helps to utilize the model capacity efficiently to learn more tasks but also led to sparsity in the activation of the neurons which reduces the interference between the tasks. It is worth highlighting that our findings from this research are aligned with the early work by French \cite{french1991using}. French argued that catastrophic forgetting is a direct consequence of the representational overlap of different tasks and semi-distributed representations could reduce the catastrophic forgetting problem.

Next, we analyze how the distribution of the connections changes as a result of the adaptive training. We visualize the sparse connections of the second task of the Split MNIST benchmark before and after its training. The initially allocated connections are randomly distributed between the selected neurons as shown in Figure \ref{initial_W}. Instead of having the sparse connections distributed over all the selected neurons, the evolution procedure makes the connections of a task grouped in a compact number of neurons as shown in Figure \ref{W_after_train}, leaving space for future tasks.

\begin{figure}
 \centering
 \begin{subfigure}[b]{0.3\columnwidth}
     \centering
     \includegraphics[width=0.97\columnwidth]{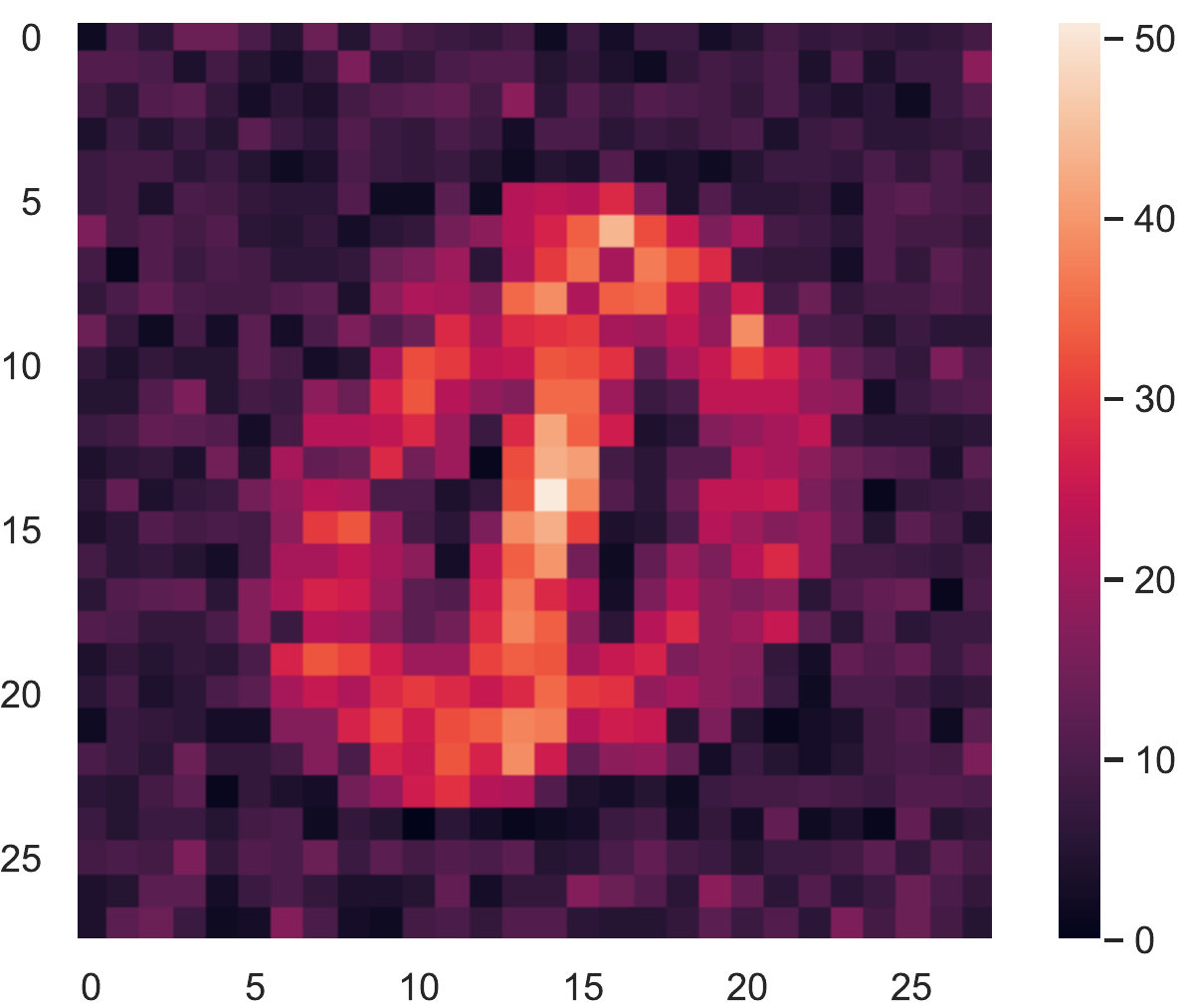}
     \includegraphics[width=0.97\columnwidth]{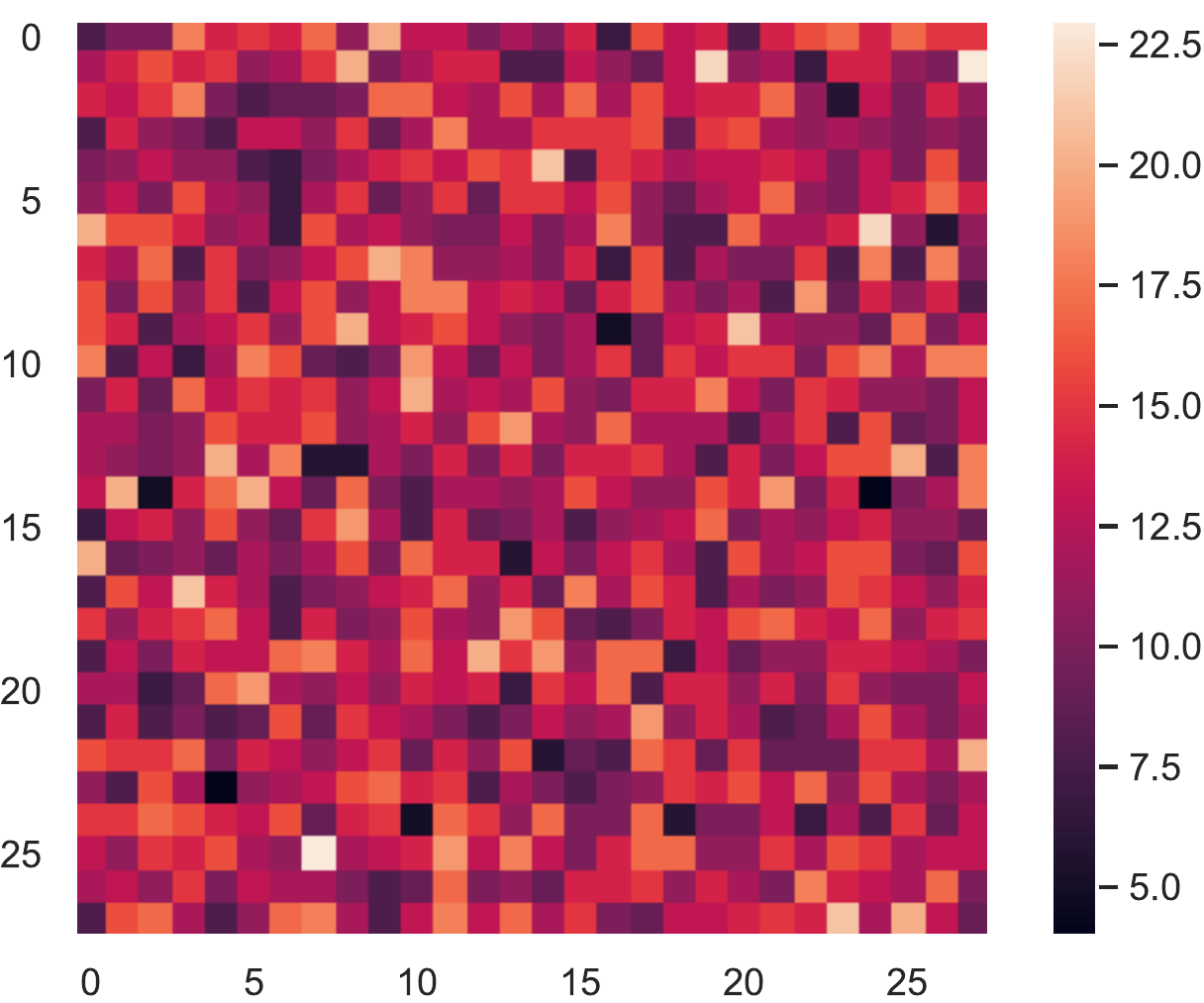}
     \caption{Task 1 (0 or 1).}
     \label{connectionsTask1}
\end{subfigure}
 \begin{subfigure}[b]{0.3\columnwidth}
    \centering
    \includegraphics[width=0.97\columnwidth]{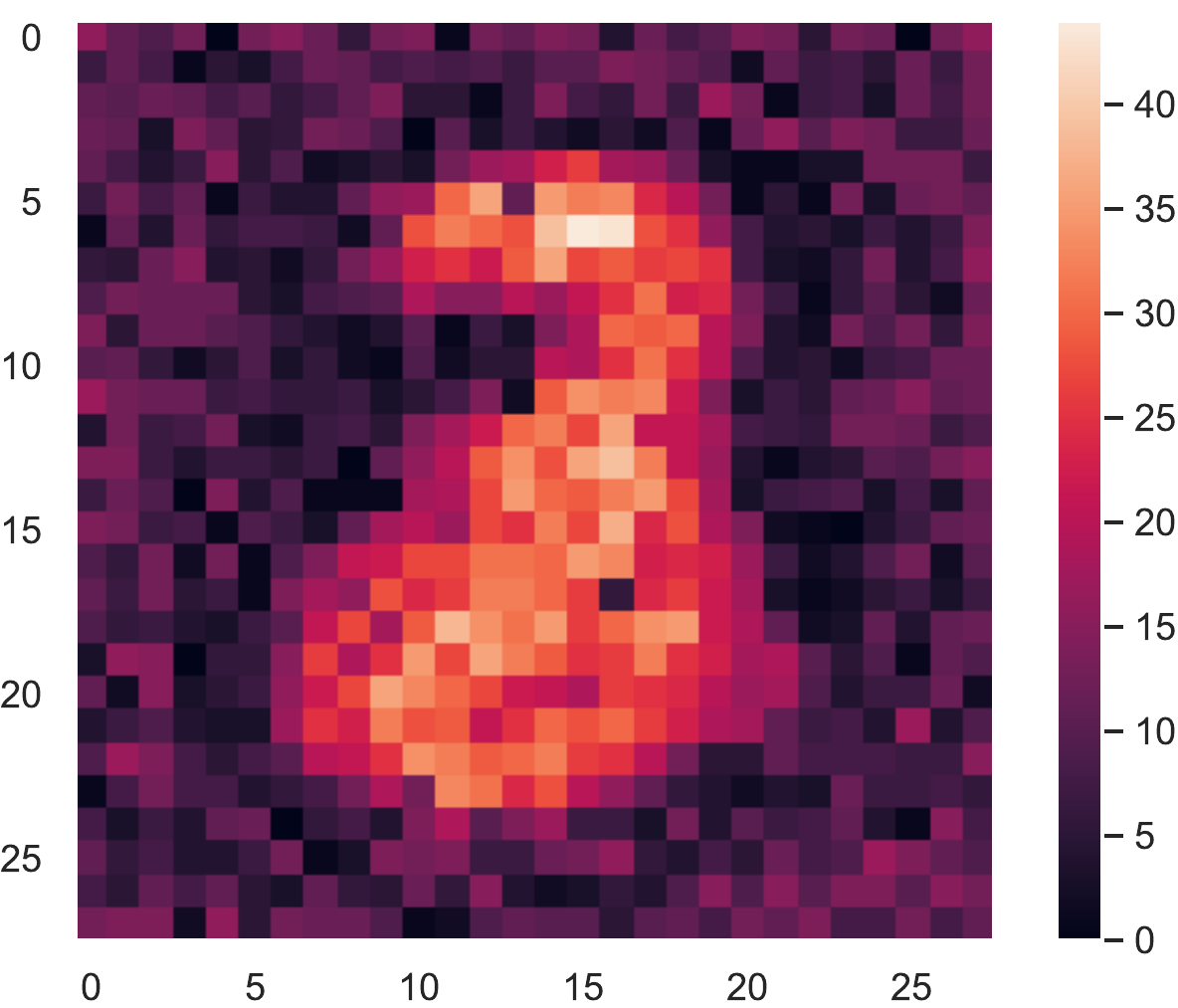}
    \includegraphics[width=0.97\columnwidth]{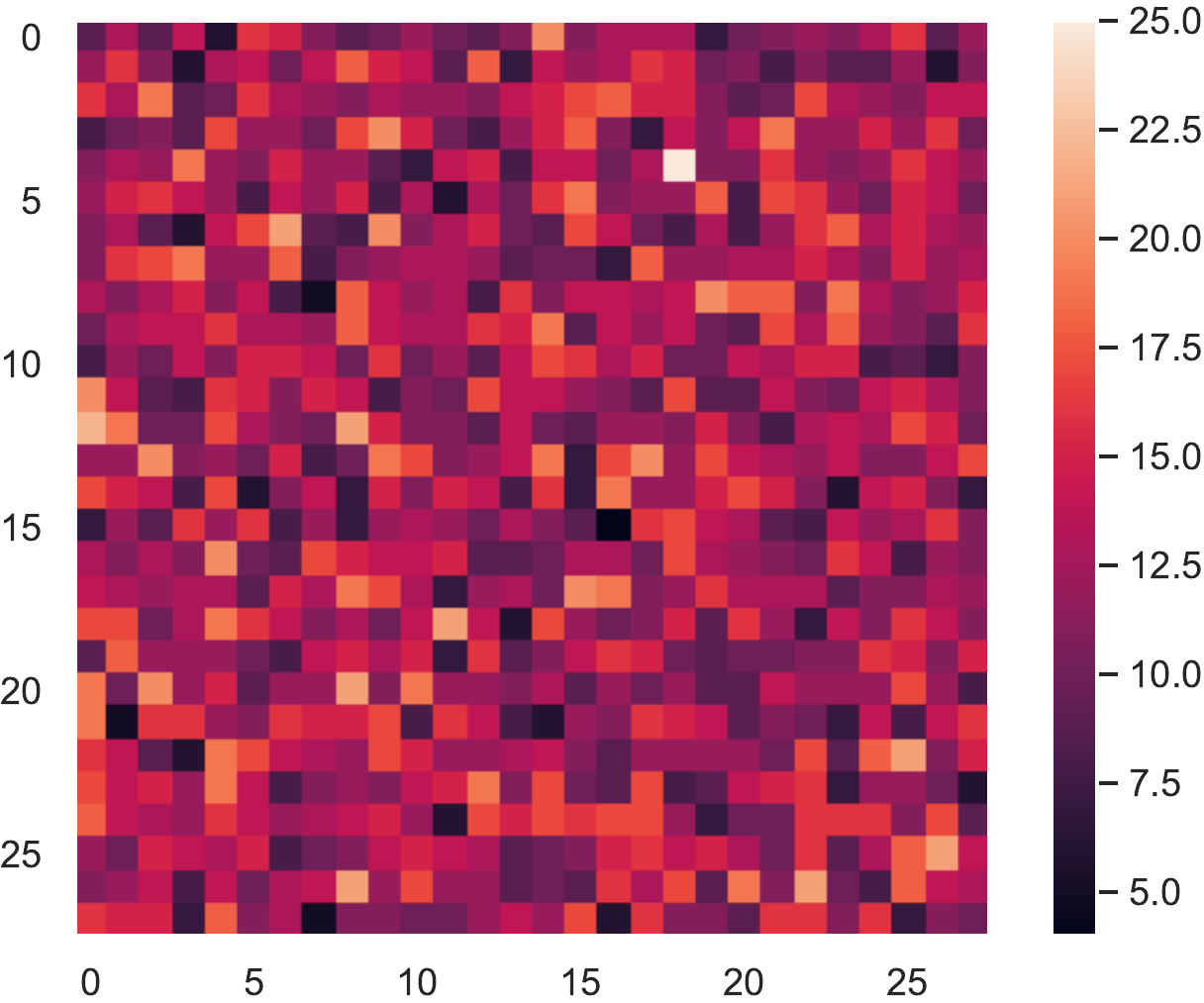}
      \caption{Task 2 (2 or 3).}
     \label{connectionsTask2}
 \end{subfigure}
  \begin{subfigure}[b]{0.3\columnwidth}
    \centering
    \includegraphics[width=\columnwidth]{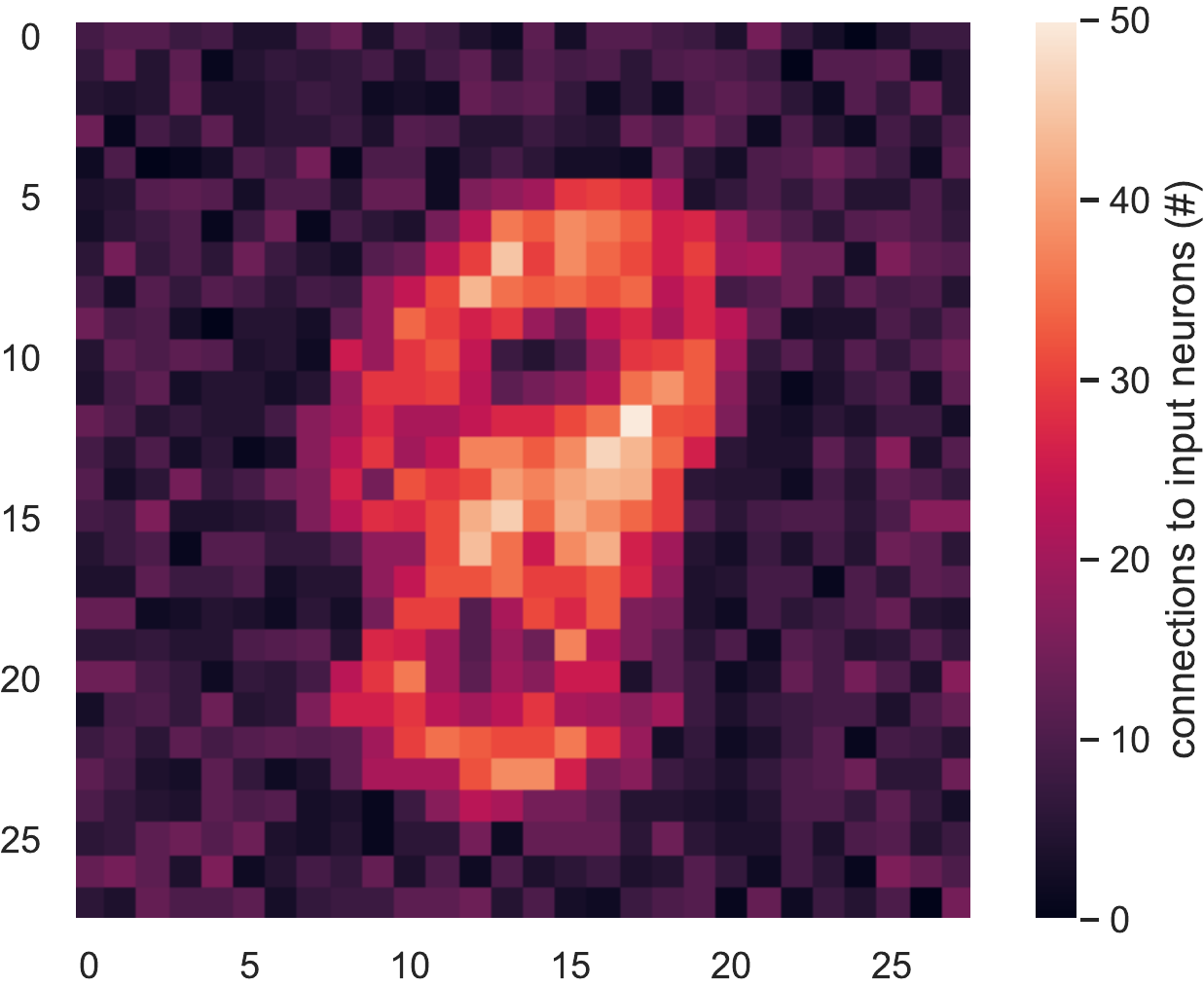}
    \includegraphics[width=\columnwidth]{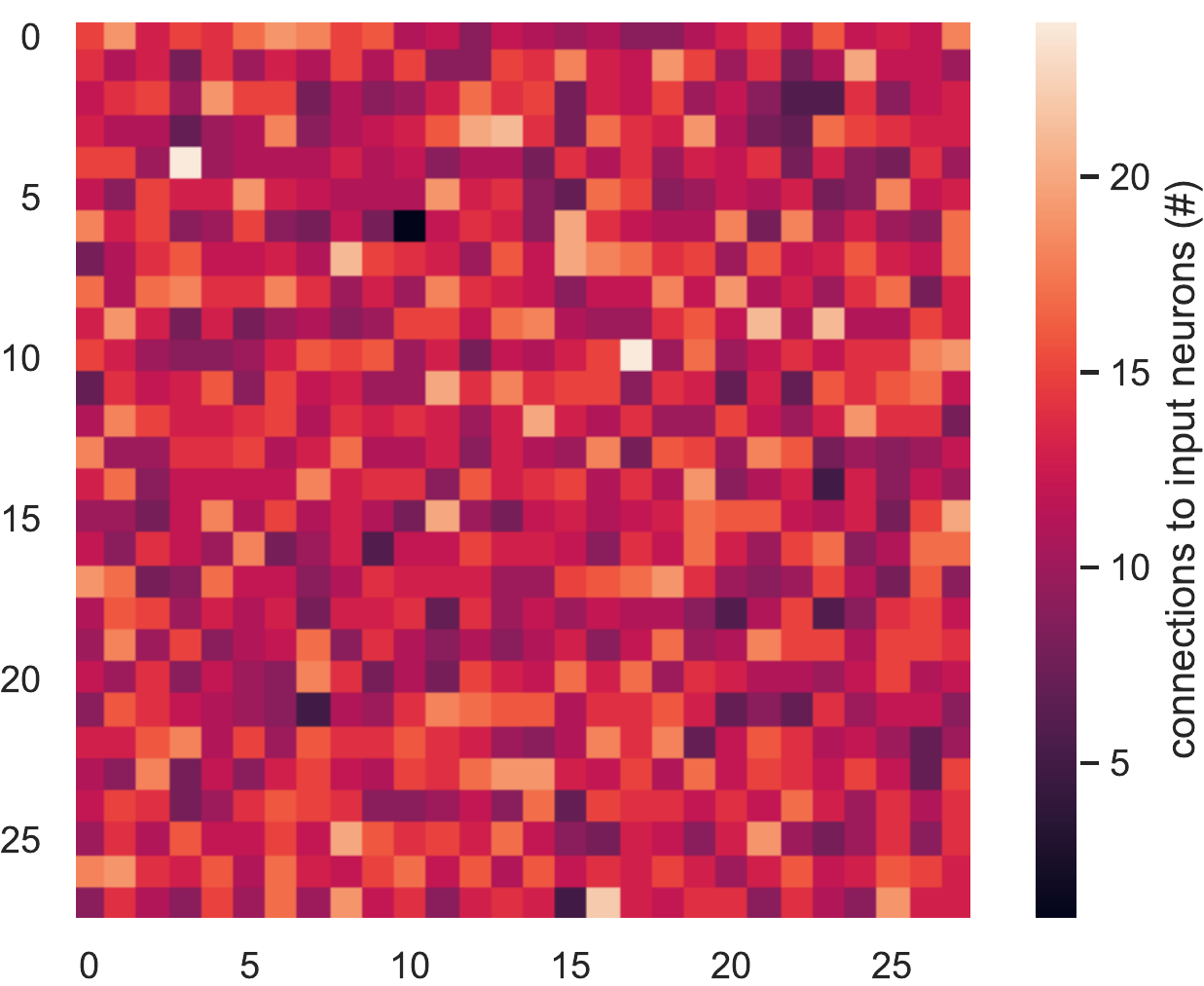}
      \caption{Task 5 (8 or 9).}
     \label{connectionsTask5}
 \end{subfigure}
 \caption{Visualization of the number of connected weights to each of the input neurons for three different tasks in the Split MNIST benchmark. The connections are reshaped to $28\times28$ to be visualized as an image. The first row represents the connections distribution results from our proposed method, SpaceNet. While the second row results from the \enquote{Static-SparseNN} baseline discuss in the experiments section.}
 \label{connections_for_each_task}
\end{figure}

We further analyze whether the connections are grouped in the right neurons (e.g. the important ones) or not. To qualitatively evaluate this point, we visualize the number of existing connections outgoing from each neuron in the input layer. The input layer consists of 784 neurons ($28\times28$). Consider the first layer of the multilayer perception network used for the Split MNIST benchmark. The layer is parameterized by the sparse weights $W_{l=1} \in R^{784 \times 400}$. We visualize the learned connections corresponding to some of Split MNIST tasks. For each $W_{l=1}^{t}$, we sum over each row to get the number of connections linked to each of the 784 input neurons. We then reshape the output vector to $28\times28$. Figure \ref{connections_for_each_task} shows the visualization of connections distribution for three different tasks of the Split MNIST benchmark. As shown in the figure, more connections are grouped in the input neurons that define the shape of each digit.
 For example in Figure \ref{connectionsTask1}, in the first row, most of the connections are grouped in the neurons representing class 0 and class 1. The figure also illustrates the distribution of the connections in the case of \enquote{Static-SparseNN} baseline discussed in the experiments section. As shown in the figure, in the second row, the connections are distributed over all the neurons of the input layer regardless of the importance of this neuron to the task which could lead to the interference between the tasks.  

\section{Conclusion}
In this work, we have proposed SpaceNet, a new technique for deep neural networks to learn a sequence of tasks in the continual learning paradigm. SpaceNet learns each task in a compact space in the model with a small number of connections, leaving a space for other tasks to be learned by the network. We address the class incremental learning scenario, where the task identity is unknown during inference. The proposed method is evaluated on the well-known benchmarks for CL: split MNIST, split Fashion-MNIST, CIFAR-10/100, and iCIFAR100. 

Experimental results show the effectiveness of SpaceNet in alleviating the catastrophic forgetting problem. Results on split MNIST and split Fashion-MNIST outperform the existing well-known regularization methods by a big margin: around 51\% and 44\% higher accuracy on the two datasets respectively, thanks to the technical novelty of the paper. SpaceNet achieved better performance than the existing architectural methods, while using a fixed model capacity without network expansion. Moreover, the accuracy of SpaceNet is comparable to the studied rehearsal methods and satisfactory given that we use 28 times lower memory footprint and do not use the old tasks data during learning new tasks. It worths mentioning that even if it was a bit outside of the scope of this paper, when we combined SpaceNet with a rehearsal strategy, the hybrid obtained method (i.e. SpaceNet-Rehearsal) outperformed all the other methods in terms of accuracy.
The experiments also show how the proposed method efficiently utilizes the available space in a small CNN architecture to learn a sequence of tasks from more complex benchmarks: CIFAR-10/100 and iCIFAR100. Unlike other methods that have a high performance on the last learned task only, SpaceNet is able to maintain good performance on previous tasks as well. Its average accuracy computed overall tasks is higher than the ones obtained by the state-of-the-art methods, while the standard deviation is much smaller. This demonstrates that SpaceNet has the best trade-off between non-catastrophic forgetting and using a fixed model capacity.  

The proposed method showed its success in addressing more desiderata for CL besides alleviating the catastrophic forgetting problem such as: memory efficiency, using a fixed model size, avoiding any extra computation for adding or retaining knowledge, and handling the inaccessibility to old tasks data. We finally showed that the learned representations by SpaceNet is highly sparse and the adaptive sparse training results in redistributing the sparse connections in the important neurons for each task.

There are several potential research directions to expand this work. In the future, we would like to combine SpaceNet with a resource-efficient generative-replay method to enhance its performance in terms of accuracy, while reducing even more the memory requirements. Another interesting direction is to investigate the effect of balancing the magnitudes of the weights across all tasks to mitigate the bias towards a certain task.

\balance
\bibliographystyle{cas-model2-names}

\bibliography{cas-refs}


\end{document}